\documentclass[10pt,journal,compsoc]{IEEEtran}

\ifCLASSOPTIONcompsoc
  \usepackage[nocompress]{cite}
\else
  \usepackage{cite}
\fi
\ifCLASSINFOpdf
\else
\fi
\usepackage{bm}
\usepackage{microtype}

\usepackage[T1]{fontenc}    
\usepackage[dvipsnames]{xcolor}
\usepackage{graphicx}
\usepackage{subfigure}
\usepackage{booktabs} 
\usepackage{multirow}
\usepackage{amssymb}
\usepackage{algorithmic}
\usepackage[ruled,linesnumbered]{algorithm2e}
\definecolor{lightseagreen}{rgb}{0.13, 0.7, 0.67}
\definecolor{highlight}{HTML}{355C7D}

\usepackage{pifont}

\usepackage{amsthm}
\usepackage{amsmath} 
\usepackage{ragged2e}

\usepackage{subfiles}
\usepackage{caption}
\usepackage{makecell}
\usepackage{graphicx}

\usepackage{url}
\usepackage{enumitem}

\newcommand{\PreserveBackslash}[1]{\let\temp=\\#1\let\\=\temp}

\def \fk#1 {\mathcal{F}(\boldsymbol{W}^{#1},\boldsymbol{U}^{#1})}
\def \fkt#1 {\widetilde{\mathcal{F}}(\boldsymbol{W}^{#1},\boldsymbol{U}^{#1})}
\def \lk#1 {\mathcal{J}(\boldsymbol{W}^{#1})}

\def \ctp1{\bm{\varTheta}_c^{t+1} }
\def \gtp1{\bm{\varTheta}_g^{t+1} }
\def \ptp1{\bm{\varTheta}_p^{t+1} }

\AtBeginDocument{%
	\providecommand\BibTeX{{%
			\normalfont B\kern-0.5em{\scshape i\kern-0.25em b}\kern-0.8em\TeX}}}

\usepackage{bbm}

\usepackage[framemethod=tikz]{mdframed}
\usepackage{lipsum}

\newtheorem{Assumption}{Assumption}
\newtheorem{Proposition}{Proposition}

\begin{document}
%
\title{Learning on Attribute-Missing Graphs}
%
%
%
%

\author{Xu~Chen, Siheng~Chen, Jiangchao~Yao, Huangjie~Zheng, Ya~Zhang*, and~Ivor~W~Tsang
\IEEEcompsocitemizethanks{
\IEEEcompsocthanksitem Xu Chen is with the Cooperative Medianet Innovation Center and the Shanghai Key Laboratory of Multimedia Processing and Transmissions, Shanghai
Jiao Tong University, Shanghai 200240, China. He is also with the Australian Artificial Intelligence Institute (AAII) at University of Technology Sydney, 15 Broadway, Ultimo NSW 2007, Sydney, Australia.  (email: \texttt{xuchen2016@sjtu.edu.cn}).\protect\\
\IEEEcompsocthanksitem Siheng Chen is with Mitsubishi Electric Research Laboratories, Cambridge, MA 02139, USA. (email: \texttt{schen@merl.com}).\protect\\
\IEEEcompsocthanksitem Jiangchao Yao is with the Alibaba Group, Hangzhou, China. (email: \texttt{jiangchao.yjc@alibaba-inc.com}).\protect\\
\IEEEcompsocthanksitem Huangjie Zheng is with the Texas University, Austin, US. (e-mail: \texttt{huangjie.zheng@utexas.edu}).\protect\\
\IEEEcompsocthanksitem Ya Zhang is with the Cooperative Medianet Innovation Center and the Shanghai Key Laboratory of Multimedia Processing and Transmissions, Shanghai Jiao Tong University, Shanghai 200240, China. Prof. Ya Zhang is also the corresponding author.  (e-mail: \texttt{ya\_zhang@sjtu.edu.cn}).\protect\\
\IEEEcompsocthanksitem Ivor W Tsang is the research director at University of Technology Sydney, 15 Broadway, Ultimo NSW 2007, Sydney, Australia. He is also a future fellow, a core member of Australian Artificial Intelligence Institute (AAII).  (e-mail: \texttt{ivor.tsang@uts.edu.au}).\protect\\
\IEEEcompsocthanksitem This work is supported by the National Key Research and Development Program of China (No. 2019YFB1804304), SHEITC (No. 2018-RGZN-02046), 111 plan (No. BP0719010), and STCSM (No. 18DZ2270700), and State Key Laboratory of UHD Video and Audio Production and Presentation. This work is also supported by ARC DP180100106 and DP200101328 of Australia.
   }
   }
%
%

\markboth{To Appear in IEEE TRANSACTIONS ON PATTERN ANALYSIS AND MACHINE INTELLIGENCE}%
{Chen \MakeLowercase{\textit{et al.}}: Bare Demo of IEEEtran.cls for Computer Society Journals}
%



\IEEEtitleabstractindextext{%
\begin{abstract}
\justifying
Graphs with complete node attributes have been widely explored recently. While in practice, there is a graph where attributes of only partial nodes could be available and those of the others might be entirely missing. This attribute-missing graph is related to numerous real-world applications and there are limited studies investigating the corresponding learning problems. Existing graph learning methods including the popular GNN cannot provide satisfied learning performance since they are not specified for attribute-missing graphs. Thereby, designing a new GNN for these graphs is a burning issue to the graph learning community. In this paper, we make a \emph{shared-latent space} assumption on graphs and develop a novel distribution matching based GNN called structure-attribute transformer (SAT) for attribute-missing graphs. SAT leverages structures and attributes in a decoupled scheme and achieves the joint distribution modeling of structures and attributes by distribution matching techniques.
It could not only perform the link prediction task but also the newly introduced \emph{node attribute completion} task. Furthermore, practical measures are introduced to quantify the performance of \emph{node attribute completion}. 
Extensive experiments on seven real-world datasets indicate SAT shows better performance than other methods on both link prediction and \emph{node attribute completion} tasks. Codes and data are available online:~\url{https://github.com/xuChenSJTU/SAT-master-online}
\end{abstract}

}

\maketitle

\IEEEdisplaynontitleabstractindextext

%
\IEEEpeerreviewmaketitle

\IEEEraisesectionheading{\section{Introduction}\label{sec:introduction}}

%
%
%
%

\begin{figure}[t]
\centering
\begin{minipage}[t]{0.5\textwidth}
\centering
\includegraphics[width=\textwidth]{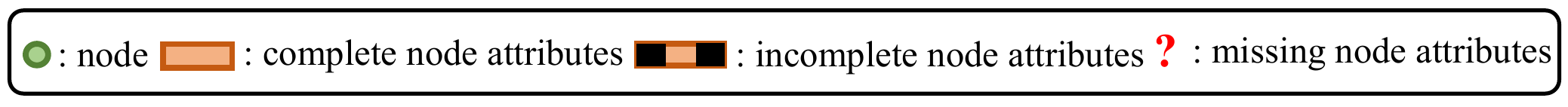}
\end{minipage} \\
\vspace{0pt}
\begin{minipage}[t]{0.15\textwidth}
\centering
\includegraphics[width=\textwidth]{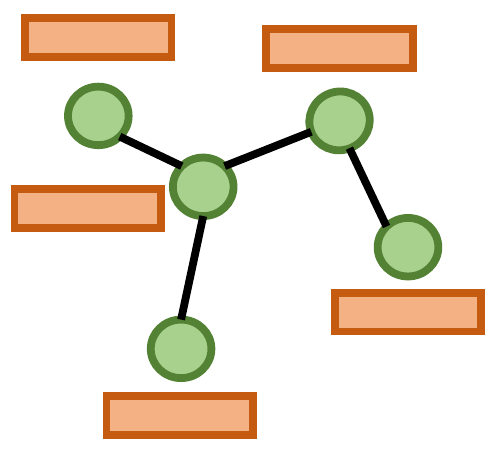}
\caption*{(a)}
\end{minipage}
\hspace{5pt}
\begin{minipage}[t]{0.15\textwidth}
\centering
\includegraphics[width=\textwidth]{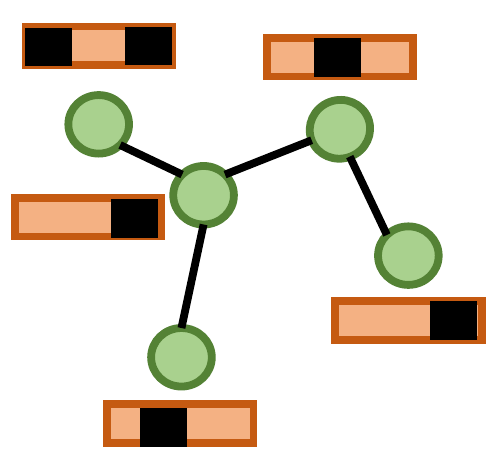}
\caption*{(b)}
\end{minipage}
\begin{minipage}[t]{0.15\textwidth}
\centering
\includegraphics[width=\textwidth]{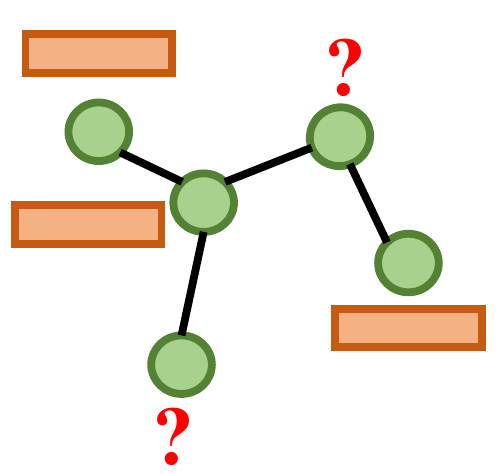}
\caption*{(c)}
\end{minipage}
\caption{Given a graph with attributes, based on the completeness of the node attributes, we may classify the graph into three types. (a) Attribute-complete graph; (b) Attribute-incomplete graph; (c) Attribute-missing graph.}
\label{figure:scenario_classification}
\end{figure}

\IEEEPARstart {G}{r}aphs, an important type of data structure, are widely employed to model citation networks, social networks, molecular graph structures, and etc. Graph-structured data are heterogeneous in nature, with correlated information about graph structures and node attributes. Given a graph with attributes, based on the completeness of the node attributes, we may classify the graph into the following three types. 1) Attribute-complete graph: every node is with complete set of attributes; 2) Attribute-incomplete graph : every node is with a non-empty set of attributes; 3) Attribute-missing graph : attributes of partial nodes are entirely missing. These three scenarios are illustrated in Figure~\ref{figure:scenario_classification}. We here mainly focus on the third one, learning with which is the most challenging among the three. The attribute-missing one is related to real-world applications. For example, in citation networks, raw attributes or detailed descriptions of some papers may be inaccessible due to copyright protection. In item co-occurrence graphs, expressive tags of some items may be unavailable because of limited tagging resources. 

\begin{figure*}[t]
\centering
\begin{minipage}[t]{0.25\textwidth}
\centering
\includegraphics[width=\textwidth]{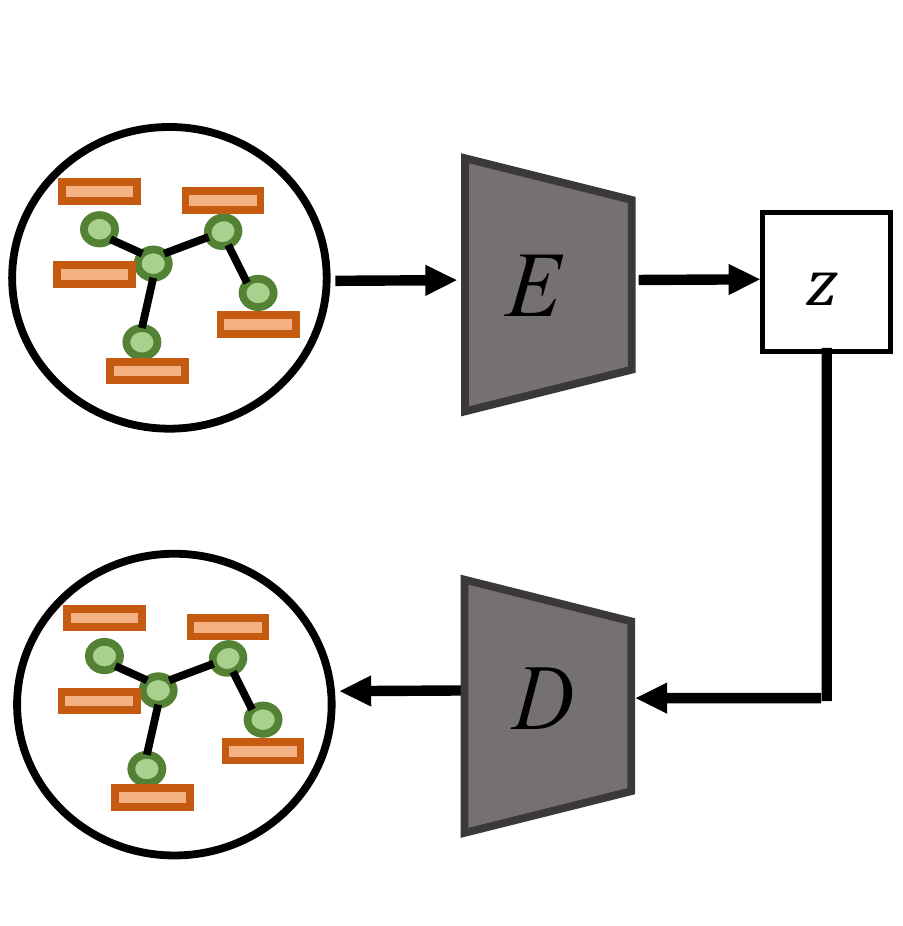}
\caption*{(a) Recent GNN}
\end{minipage}
\hspace{5pt}
\begin{minipage}[t]{0.25\textwidth}
\centering
\includegraphics[width=\textwidth]{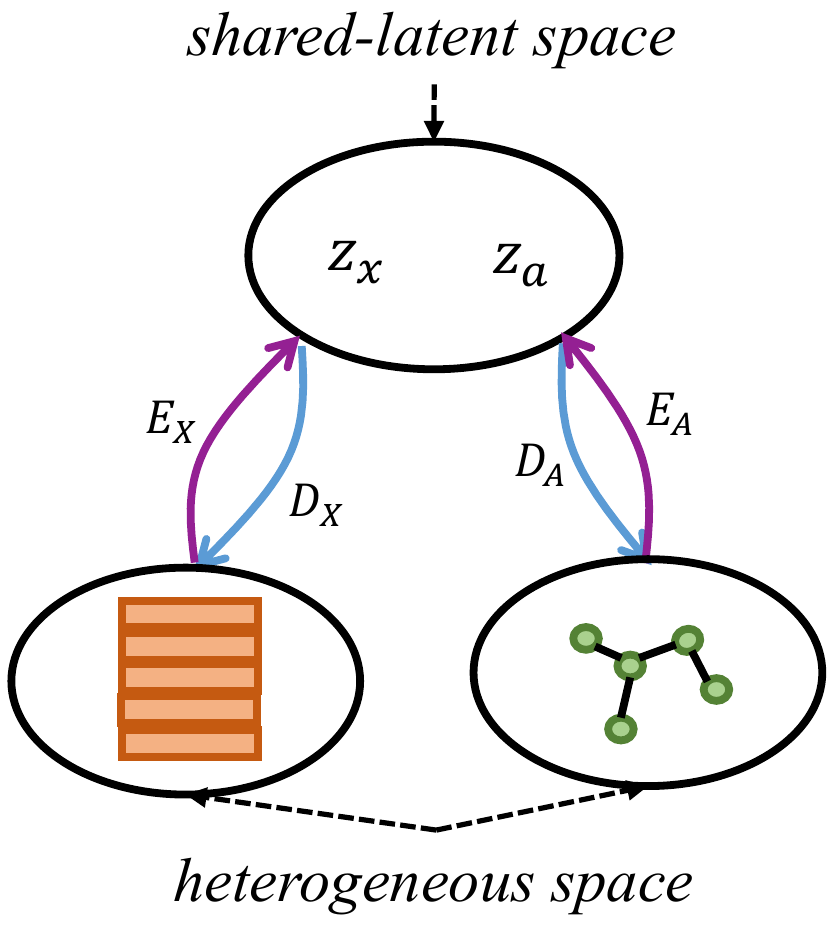}
\caption*{(b) Our \emph{shared-latent space} assumption}
\end{minipage}
\begin{minipage}[t]{0.4\textwidth}
\centering
\includegraphics[width=\textwidth]{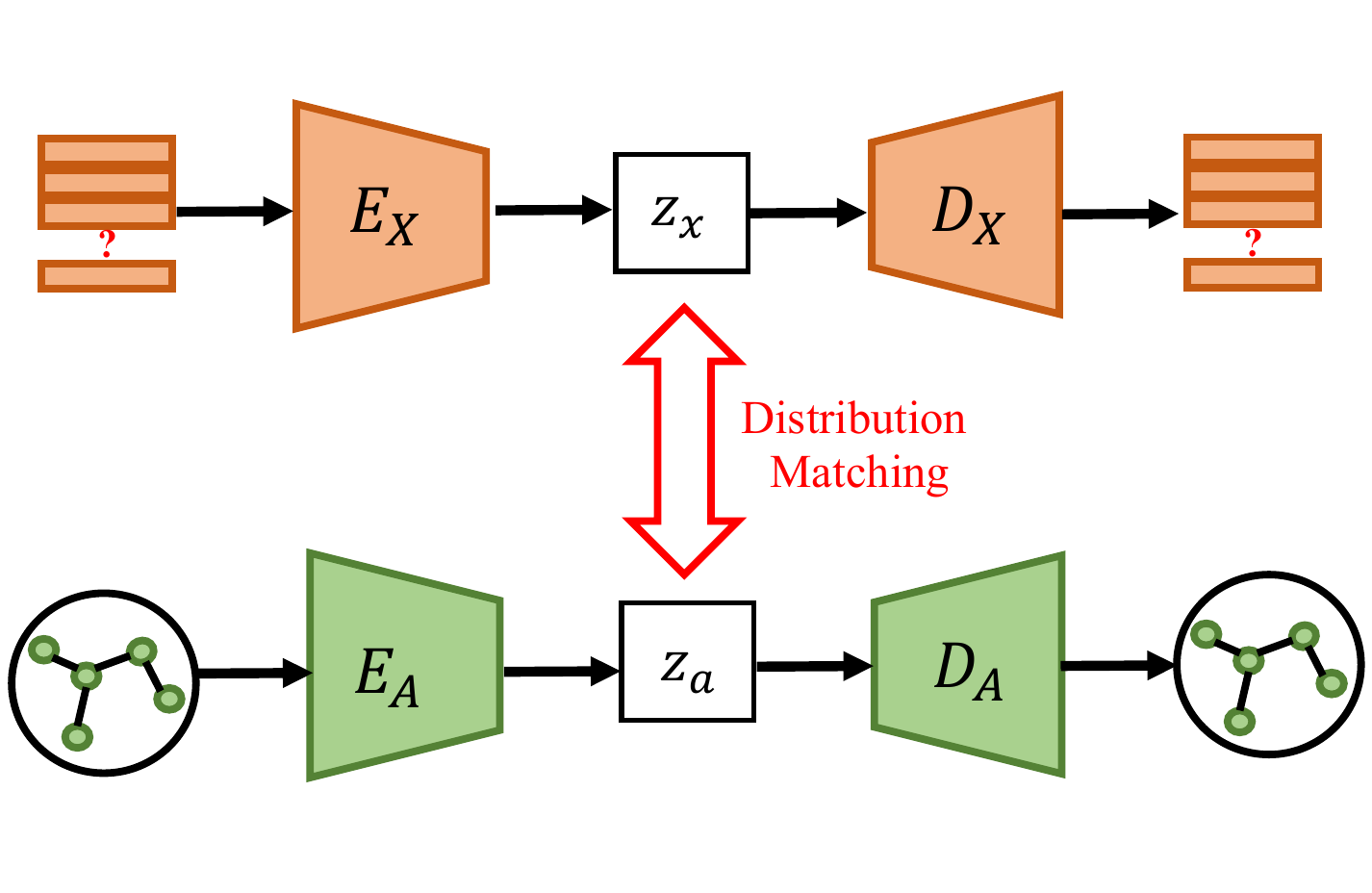}
\caption*{(c) Our SAT}
\end{minipage}
\caption{The comparison between recent GNN and our SAT, together with the illustration of our \emph{shared-latent space} assumption. (a)means recent GNN usually requires structures and attributes as a whole input. $E$ is the encoder and $D$ is the decoder. (b) shows our \emph{shared-latent space} assumption where $E_{X}$ and $E_{A}$ are two encoders and $D_{X}$ and $D_{A}$ are two decoders. (c) shows the general architecture of our SAT.}
\label{figure:GNN_comparison}
\end{figure*}
Learning on attribute-incomplete graphs can be solved by popular matrix factorization based methods~\cite{mnih2008probabilistic,monti2017geometric,berg2017graph}.
However, there are limited studies investigating learning on attribute-missing graphs. Existing graph learning methods such as the random walk based~\cite{perozzi2014deepwalk,grover2016node2vec}, the attributed random walk based~\cite{huang2019graph,chen2019} and the graph neural network (GNN)~\cite{kipf2016semi,kipf2016variational,velivckovic2017graph,hamilton2017inductive,wu2019simplifying} are not specified for attribute-missing graphs and are limited in solving the corresponding learning problems. Random walk based methods emerged as large-scale and effective graph embedding approaches which only take structures into consideration. They cannot take the advantage of rich information from node attributes~\cite{kipf2016semi}. Although the attributed random walk based methods~\cite{huang2019graph,chen2019} can potentially deal with the attribute-missing graph, these methods rely on high-quality random walks and require carefully designed sampling strategies 
and fine-tuned hyper-parameters~\cite{yang2019advances}.
GNN leverages structures and attributes as a whole in a unified framework as shown in Figure~\ref{figure:GNN_comparison} (a) and has achieved superior performance on various graph mining tasks~\cite{kipf2016semi,kipf2016variational,ying2018graph,ying2018hierarchical}.
On attribute-missing graphs with only partial nodes associated with attributes, GNN can work with some attribute-filling tricks such as the zero-filling. However, these tricks would introduce noise to the learning process.    
Therefore, designing a specific GNN method for attribute-missing graphs is a burning issue to the graph learning community.

One possible way to solve the issue is to input structures and attributes in a decoupled scheme and simultaneously allow the joint distribution modeling of structures and attributes. Structures and attributes are two resources that come from two marginal distributions. The coupling theory~\cite{lindvall2002lectures} states that there is an infinite set of joint distributions that can reach the given marginal distributions in general. Therefore, it is impossible for us to perform the joint distribution modeling of structures and attributes without any assumption. Besides, the graph structured data has two characteristics: 1) the structures and attributes come from two heterogeneous spaces (\emph{heterogeneity}); 2) the node attributes can either be real-valued or categorical (\emph{discontinuity}). These two characteristics hinder us from achieving joint distribution modeling in data space by existing techniques such as weight-sharing in the encoder and adversarial learning in data space~\cite{tzeng2017adversarial,makhzani2015adversarial,isola2017image}. To this end, we make a \emph{share-latent-space assumption} on graphs which assumes that the heterogeneous structures and attributes are related to each other and come from the same latent space (illustrated in Figure~\ref{figure:GNN_comparison} (b)).

In this paper, we achieve the \emph{shared-latent space} assumption by distribution matching techniques and further develop a novel distribution matching based GNN for learning on attribute-missing graphs, called structure-attribute transformer (SAT). The general architecture of SAT is shown in Figure~\ref{figure:GNN_comparison} (c).
SAT leverages structures and attributes in a decoupled scheme and achieves the joint distribution modeling by matching the latent codes of structures and attributes. It can not only perform the link prediction task but also retrieve the missing attributes. We also call the latter as \emph{node attribute completion} which benefits several subsequent tasks such as node classification and profiling.
The \emph{node attribute completion} is a new problem on graphs which distinguishes from other existing problems and requires specific evaluation measures. Main contributions of this paper can be summarized as follows:
\begin{itemize}
 \item We investigate the learning problems on attribute-missing graphs. In this scenario, we make a~\emph{shared-latent space} assumption on graphs and develop a novel distribution matching based GNN called SAT. SAT is a generic framework that can be equipped with other popular GNN backbones such as GCN and GAT;
 \item We introduce a new task called \emph{node attribute completion} on graphs which aims to complete the missing node attributes and benefit other subsequent tasks. To quantify the performance of \emph{node attribute completion}, practical measures are introduced including both node classification in the \emph{node level} and profiling in the \emph{attribute level};
 \item Extensive experiments on seven real-world datasets show that our method can not only perform well on link prediction task but also restore high-quality attributes that benefit subsequent tasks such as node classification and profiling.
\end{itemize}

\section{Related Work}
\subsection{Deep Learning on Graphs}
With the success of deep neural networks in modeling speech signals, images and texts, many researchers start to apply deep neural networks to the graph domain. Recent deep learning on graphs mainly concentrates on three perspectives: the random walk based, the attributed random walk based and the GNN based methods.  
The random walk based methods such as DeepWalk~\cite{perozzi2014deepwalk} and Node2Vec~\cite{grover2016node2vec} learn node embeddings by random walks and skip-gram model~\cite{mikolov2013efficient}. They only take graph structures into consideration and cannot exploit the rich information from node attributes. Hence, the attributed random walk based methods~\cite{huang2019graph,chen2019} emerged to tackle this problem. GraphRNA~\cite{huang2019graph} takes node attributes as a bipartite graph and use it to obtain diverse random walks. Then, network embedding approaches in previous methods~~\cite{perozzi2014deepwalk,grover2016node2vec,taheri2018learning} can be employed. The attributed random walk based methods are summarized as a matrix factorization formulation in~\cite{chen2019}. These methods rely much on high-quality random walks and are labour-consuming for carefully designed sampling strategies and fine-tuned hyper-parameters. 

Graph convolution networks (GCN) is proposed in~\cite{defferrard2016convolutional,kipf2016semi}, and it is successfully applied in semi-supervised node classification problem. A hierarchical and differentiable pooling is proposed in~\cite{ying2018hierarchical,lee2019self} to learn representation for graphs. GAT~\cite{velivckovic2017graph} applies multi-head attention to graph neural networks. GraphSAGE~\cite{hamilton2017inductive} introduces neighborhood sampling and different aggregation manners to make inductive graph convolution on large graphs. SGC in \cite{wu2019simplifying} simplifies the graph convolution operation for fast learning and low complexity. The idea of relation modeling behind GCN has been extended to various real-world applications such as recommendation~\cite{ying2018graph,wang2018ripplenet,wang2019kgat} and skeleton based action recognition~\cite{yan2018spatial,li2019actional}. 
In addition, there are some works studying GNN on the matrix completion problem. For example, Rianne et.al~\cite{berg2017graph} consider the interaction data in recommendation systems as a a bipartite user-item graph and a link prediction based GNN is proposed. Federico et.al~\cite{federico2017gcmc} proposed GCMC that combines a novel
multi-graph convolutional neural network and a recurrent neural network to better exploit the local stationary structures on the user-item graph. Note that these graph-based matrix completion methods~\cite{berg2017graph,federico2017gcmc} are not feasible on attribute-missing graphs since attributes of some nodes are completely missing. Although some advanced matrix completion approaches~\cite{hu2019hers} working on cold-start recommendation can potentially handle \emph{node attribute completion} task, they simply take the graph structures as one kind of side information and cannot well exploit the joint relationship between attributes and structures.

In many real-world applications, graph structures may be missing, incomplete or inaccessible. To handle this, graph generation has been studied recently. A junction tree variational auto-encoder is proposed in~\cite{jin2018junction} to generate molecular graphs. MolGAN~\cite{de2018molgan} is an implicit and likelihood-free generative model to generate small molecular graphs. GraphRNN~\cite{you2018graphrnn} and NetGAN~\cite{bojchevski2018netgan} generate realistic graphs by the combination of random walks and recurrent neural networks (RNN). Despite the great potential in many applications such as link prediction and \emph{node attribute completion} on attribute-missing graphs, few works investigate it and we are going to explore it in this paper. 

\subsection{Distribution Matching}
Distribution matching is a statistical technique for distribution alignment. It has been considerably discussed by various methodologies and extended to many applications. In recent years, variational inference~\cite{hoffman2013stochastic} and adversarial learning~\cite{goodfellow2014generative} are two representative techniques for distribution matching. Variational inference imposes distribution matching in a probabilistic way with specified objective functions. For example, VAE~\cite{kingma2013auto} introduces Kullback-Leibler (KL) divergence to match the latent posterior distribution with the Gaussian prior. Based on VAE, various variants~\cite{zhao2017infovae,Zheng2018DegenerationIV,Zheng2019UnderstandingVI} are proposed to improve the encoding and reconstruction performance, together with better distribution matching manners. For example, Zhao et.al~\cite{zhao2017infovae} proposed InfoVAE with mutual information maximization in the objective function. Zheng et.al~\cite{Zheng2019UnderstandingVI} investigate VAE in the Fisher-Shannon plane and propose FAE to balance Fisher information and Shannon information in VAE. Both of InfoVAE and FAE improve the quality of the variational posterior and encourage better distribution matching.

Adversarial learning in GAN~\cite{goodfellow2014generative,nowozin2016f,mao2017least,arjovsky2017wasserstein} is another hot distribution matching technique. GAN contains a generator and a discriminator, where the discriminator tries to distinguish the real samples with the fake samples and the generator tries to confuse the discriminator. Adversarial learning in GAN has the advantage of measuring the distribution distance in a more elegant way by binary classification and frees researchers from the painful practice of defining a tricky objective function. Several works~\cite{goodfellow2014generative,nowozin2016f,chen2016infogan} have pointed that the adversarial loss in GAN actually minimizes the Jensen-Shannon divergence (JSD) between the data distribution and the generator distribution. Original GAN has risk facing the vanishing gradient and mode collapse problem. To handle this, a lot of works have been developed to improve the objective function such as f-GAN~\cite{nowozin2016f}, LSGAN~\cite{mao2017least} and WGAN~\cite{arjovsky2017wasserstein}. GAN has been introduced to many research topics that require distribution matching such as domain adaptation~\cite{ganin2016domain,tzeng2017adversarial,long2018transferable}, image translation~\cite{liu2017unsupervised,zhu2017unpaired} and transfer learning~\cite{cao2018partial,zhou2019dual,isola2017image}. 

There are also some works~\cite{mescheder2017adversarial,srivastava2017veegan,rosca2018distribution} trying to combine VAE and GAN theory together to perform distribution matching. For example, adversarial variational Bayes (AVB) is proposed in~\cite{mescheder2017adversarial} to train VAE with arbitrary inference models. VEEGAN in~\cite{srivastava2017veegan} changes the matching manner from data space to latent space. Also, Rosca et.al~\cite{rosca2018distribution} point out that VAE based methods fail to match marginal distributions in both latent and visible space while GAN has the potential to overcome these limitations. Thus they develop a VAE-GAN hyprid model to improve distribution matching and generation quality. Our SAT also works in the auto-encoding Bayes and adversarial learning mechanism. Note that although combining VAE and GAN theory is not new in distribution matching community, it is rarely considered on graph structured data.

\begin{figure*}[t]
\centering
\includegraphics[width=14cm]{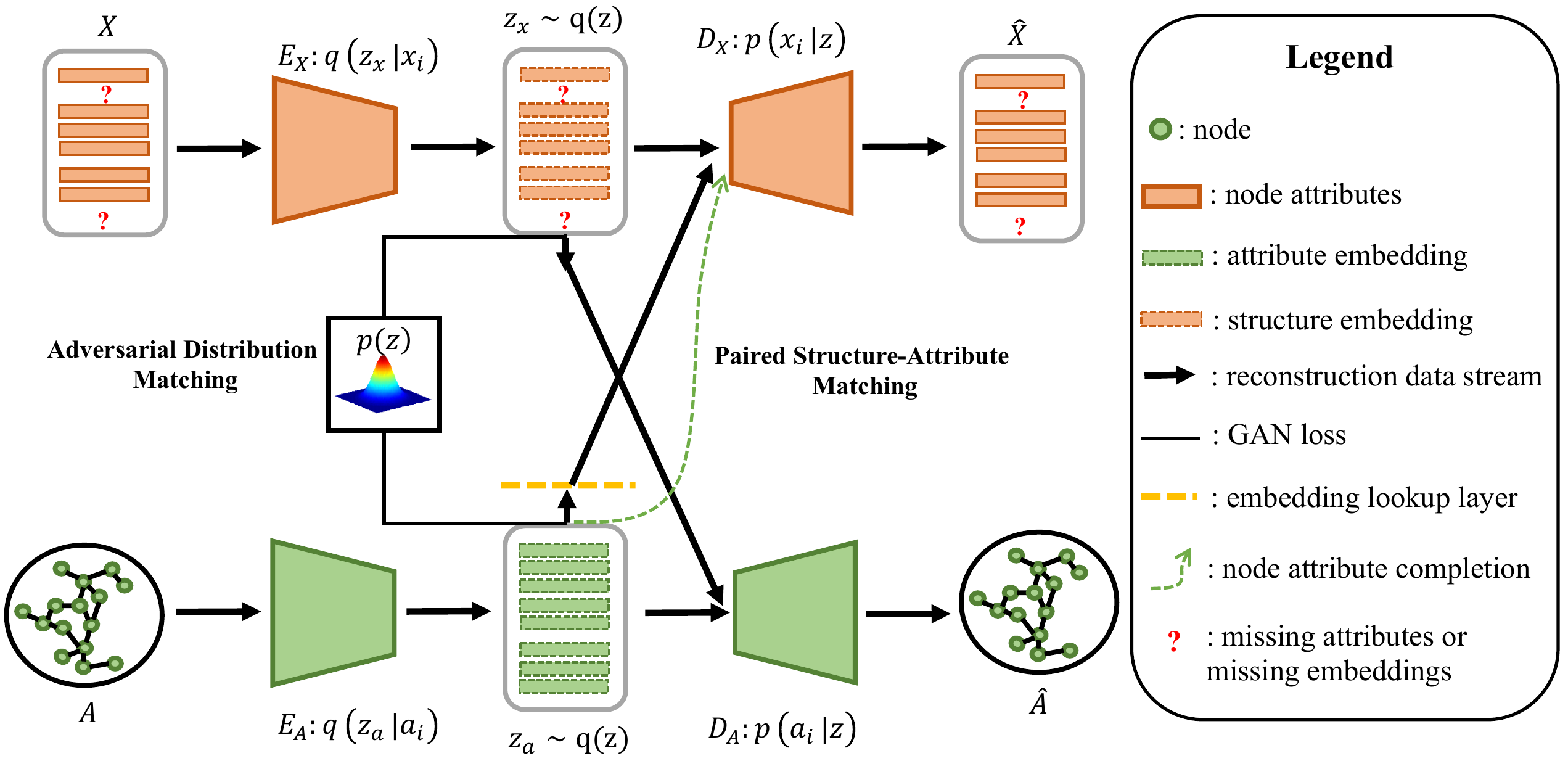}
\caption{Architecture of SAT. SAT first transforms attributes and structures into the latent space, then aligns the paired latent representations via adversarial distribution matching, and finally decodes to the original attributes and structures, namely the paired structure-attribute matching.}
\label{figure:model_architecture}
\end{figure*}

\section{Method}
\subsection{Problem Definition and Notation Description}
\label{sec:problem_definition}
Let $\mathcal{G}=(\mathcal{V},A,X)$ be a graph with node set $\mathcal{V}=\{v_{1},v_{2},...,v_{N}\}$, $A\in R^{N\times N}$ be the graph adjacent matrix and $X\in R^{N\times F}$ be the node attribute matrix.
Note that the element in $X$ could be either categorical values or real values. Let $\mathcal{V}^{o}=\{v_{1}^{o},v_{2}^{o},...,v_{N_{o}}^{o}\}$ be the set of attribute-observed nodes. Let $x_{i}^{o}\in R^{F}$ and $a_{i}^{o}$ be the attribute vector and structural information (the neighbors of a node) of node $v_{i}^{o}$, respectively. Then, for $\mathcal{V}^{o}$, the attribute information is denoted as $X^{o}=\{x_{0}^{o},x_{1}^{o},...,x_{N_{o}}^{o}\}$ and structural information is denoted as $A^{o}=\{a_{0}^{o},a_{1}^{o},...,a_{N_{o}}^{o}\}$. Let $\mathcal{V}^{u}=\{v_{1}^{u},v_{2}^{u},...,v_{N_{u}}^{u}\}$ be the set of attribute-missing nodes. Let $x_{i}^{u}\in R^{F}$ and $a_{i}^{u}$ be the attribute vector and structural information of node $v_{i}^{u}$, respectively. Then, for $\mathcal{V}^{u}$, the attribute information is denoted as $X^{u}=\{x_{0}^{u},x_{1}^{u},...,x_{N_{u}}^{u}\}$
and structural information is denoted as $A^{u}=\{a_{0}^{u},a_{1}^{u},...,a_{N_{u}}^{u}\}$. According to our definition, $A$ is another expression of $A^{o}$ and $A^{u}$. $X$ is another expression of $X^{o}$ and $X^{u}$. To clarify more clearly, we have $\mathcal{V}=\mathcal{V}^{u} \cup \mathcal{V}^{o}$, $\mathcal{V}^{u} \cap \mathcal{V}^{o} = \emptyset$ and $N = N_{o} + N_{u}$. Learning on attribute-missing graphs has several applications such as link prediction and the newly introduced \emph{node attribute completion}. In link prediction, we expect to predict the missing links in $A$. In \emph{node attribute completion}, we aim to restore the missing node attributes $X^{u}$ based on the observed node attributes $X^{o}$ and graph structures $A$. The main notations in this paper are summarized in Table~\ref{table:notations}.

\subsection{Method Formulation}
\subsubsection{Overview}
\label{section:method_formulation}
Graph-structured data include two perspectives of representations: structures and attributes. On attribute-missing graphs, attributes of some nodes might be entirely missing. 
One possible way to learn on attribute-missing graphs is to input structures and attributes in a decoupled scheme and simultaneously allow the joint distribution modeling of structures and attributes. 

\begin{table}[]
\caption{Main notations in this paper.}
\label{table:notations}
\renewcommand{\arraystretch}{1.0}
 \setlength{\tabcolsep}{0.8mm}{ 
  \scalebox{1.0}{
\begin{tabular}{ll}
\hline
Notation & Description \\ \hline
$\mathcal{G}$        & the graph           \\
$\mathcal{V}$        & the set of nodes of graph $\mathcal{G}$            \\
$A$             & the adjacent matrix of graph $\mathcal{G}$    \\
$X$            & the attribute matrix of graph $\mathcal{G}$   \\
$N$    & the number of nodes of graph $\mathcal{G}$  \\
$F$        & the attribute dimension \\
$v_{i}$        & an arbitrary node on $\mathcal{G}$ \\
$x_{i}/a_{i}$        & the attribute/structural information of node $v_{i}$ \\
$z_{x}/z_{a}$        & the latent factors of  $x_{i}/a_{i}$ \\
$z_{p}$     & samples from Gaussian prior for adversarial learning         \\
$\mathcal{V}^{o}$        & the set of attribute-observed nodes \\
$\mathcal{V}^{u}$        & the set of attribute-missing nodes \\
$X^{o}/A^{o}$        & the set of attribute/structural information for $\mathcal{V}^{o}$ \\
$X^{u}/A^{u}$        & the set of attribute/structural information for $\mathcal{V}^{u}$ \\
$d$        & the latent dimension \\
$Z_{A}$     & latent embedding encoded from structural information          \\
$Z_{X^{o}}$     & latent embedding encoded from attributes of $\mathcal{V}^{o}$          \\
$Z_{A^{o}}$     & latent embedding encoded from structures of $\mathcal{V}^{o}$          \\
$Z_{A^{u}}$     & latent embedding encoded from structures of $\mathcal{V}^{u}$           \\
$E_{X}/E_{A}$     & encoder for attribute/structural information          \\
$D_{X}/D_{A}$     & decoder for attribute/structural information          \\
$\mathcal{D}$     & shared discriminator          \\
$\mathcal{L}_{r}$    & reconstruction loss of paired structure-attribute matching  \\
$\mathcal{L}_{adv}$    & GAN loss of adversarial distribution matching  \\
$\mathcal{L}$    & objective function of SAT  \\
\hline
\end{tabular}
}}
\end{table}
In particular, if we denote $x_{i}$ and $a_{i}$ as the attribute vector and structural information of node $v_{i}$, we can see $(x_{i},a_{i})$ is a paired sample to describe node $v_{i}$. The joint log-likelihood of the attributes and structures is composed of a sum over the likelihoods of individual datapoints $\sum_{i}p_{\theta}(x_{i},a_{i})$, where $p_{\theta}(x_{i},a_{i})$ is the joint probability density function. Inspired by the idea of maximizing the marginal likelihood in VAE~\cite{kingma2013auto}, if $z_{x}$ and $z_{a}$ are the latent factors of $x_{i}$ and $a_{i}$, respectively, $\log p_{\theta}(x_{i},a_{i})$ can be formulated as:
\begin{equation}
\label{eq:original_objective}
\begin{split}
    \log p_{\theta}(x_{i},a_{i})=& D_{KL}[q_{\phi}(z_{x},z_{a}|x_{i}, a_{i})||p(z_{x},z_{a}|x_{i}, a_{i})]\\ 
    & +\mathcal{L}(\theta,\phi;x_{i},a_{i})
\end{split}
\end{equation}
where the first term is the KL divergence of the approximate posterior $q_{\phi}(z_{x},z_{a}|x_{i},a_{i})$ from the true posterior $p(z_{x},z_{a}|x_{i},a_{i})$. Since this KL term is non-negative, the second term is the \textit{evidence lower bound} (\textit{ELBO}) on the log-likelihood $\log p_{\theta}(x_{i},a_{i})$. And $\mathcal{L}(\theta,\phi;x_{i},a_{i})$ is written as:
\begin{align}
\label{eq:ELBO}
\mathcal{L}(\theta,\phi;x_{i},a_{i})=&\mathbb{E}_{q_{\phi}(z_{x},z_{a}|x_{i},a_{i})}[\log p_{\theta}(x_{i},a_{i}|z_{x},z_{a})] \nonumber \\
&-D_{KL}[q_{\phi}(z_{x},z_{a}|x_{i},a_{i})||p(z_{x},z_{a})]
\end{align}
where $p_{\theta}(x_{i},a_{i}|z_{x},z_{a})$ denotes the conditional distribution parameterized by $\theta$. The first term in Eq.~\ref{eq:ELBO} indicates the joint reconstruction loss where $z_{x},z_{a}$ encoded from $x_{i},a_{i}$ are used to reconstruct $x_{i},a_{i}$.
The second term in Eq.~\ref{eq:ELBO} indicates the prior regularization loss where $q_{\phi}(z_{x},z_{a}|x_{i},a_{i})$ is expected to match the prior distribution $p(z_{x},z_{a})$. The proposed SAT implements the the joint reconstruction loss via a paired structure-attribute matching strategy and approximates and the prior regularization loss via an adversarial distribution matching strategy. The architecture of SAT is shown in Figure~\ref{figure:model_architecture}.
In the following sections, we provide details on the two strategies, followed by the objective function and implementation.

\subsubsection{Paired Structure-Attribute Matching}
Term $\mathbb{E}_{q_{\phi}(z_{x},z_{a}|x_{i},a_{i})}[\log p_{\theta}(x_{i},a_{i}|z_{x},z_{a})]$ in Eq.~\ref{eq:ELBO} consists of an approximate posterior $q_{\phi}(z_{x},z_{a}|x_{i},a_{i})$ parameterized by $\phi$ and a conditional distribution $p_{\theta}(x_{i},a_{i}|z_{x},z_{a})$ parameterized by $\theta$. According to the chain rule~\cite{bishop2006pattern}, we have:
\begin{equation}
\label{eq:q_phi}
\begin{split}
    q_{\phi}(z_{x},z_{a}|x_{i},a_{i})= q_{\phi}(z_{x}|x_{i},a_{i})q_{\phi}(z_{a}|z_{x},x_{i},a_{i})
\end{split}
\end{equation}
Similarly, we rewrite $p_{\theta}(x_{i},a_{i}|z_{x},z_{a})$ as:
\begin{equation}
\label{eq:original_conditional}
    p_{\theta}(x_{i},a_{i}|z_{x},z_{a})=p_{\theta}(x_{i}|z_{x},z_{a})p_{\theta}(a_{i}|x_{i},z_{x},z_{a})
\end{equation}
However, the coupling theory~\cite{lindvall2002lectures} states that there exists infinite joint distribution formulations that can reach the given marginal distributions. It hinders us to derive the above joint distribution.
Moreover, the \emph{heterogeneity} and \emph{discontinuity} of graph structured data hinder us from achieving joint distribution modeling by several existing techniques such as weight-sharing in the encoder and adversarial learning in data space~\cite{tzeng2017adversarial,makhzani2015adversarial,isola2017image}. Inspired by the idea of modeling correlated data in the same latent space in recent machine learning works~\cite{liu2017unsupervised,zhu2017unpaired,wang2018deep,li2018domain}, we make the following \emph{shared-latent space} assumption on graphs, which allows us to perform the joint distribution modeling in latent space. 
\begin{Assumption}
\label{assumption:1}
\vspace{-3pt}
\textit{In graph structured data, each node's attributes and structures are correlated together and can be represented in a shared-latent space.}
\vspace{-3pt}
\end{Assumption}
This assumption is also illustrated in Figure~\ref{figure:GNN_comparison} (b). 
With this assumption, we have the following proposition:
\begin{Proposition}
\vspace{-3pt}
\label{proposition:1}
\textit{Under the shared-latent space assumption, given observations $x_{i}$, $a_{i}$, the latent variables $z_{x}$ and $z_{a}$ are conditional independent. Given latent variables $z_{x}$, $z_{a}$, the observations $x_{i}$ and $a_{i}$ are conditional independent.}
\vspace{-3pt}
\end{Proposition}
Thereby, with Proposition~\ref{proposition:1}, Eq.~\ref{eq:q_phi} can be written as:
\begin{align}
\label{eq:rewritten_q_phi}
    q_{\phi}(z_{x},z_{a}|x_{i},a_{i})&=q_{\phi_{x}}(z_{x}|x_{i},a_{i})q_{\phi_{a}}(z_{a}|x_{i},a_{i}) \nonumber \\
    &=q_{\phi_{x}}(z_{x}|x_{i})q_{\phi_{a}}(z_{a}|a_{i})
\end{align}
where $q_{\phi_{x}}(z_{x}|x_{i})$ indicates the encoder $E_{X}$ that encodes $x_{i}$ to $z_{x}$. Similarly, $q_{\phi_{a}}(z_{a}|a_{i})$ indicates the encoder $E_{A}$ that encodes $a_{i}$ to $z_{a}$. Also, together with Proposition~\ref{proposition:1} and the assumption, we write Eq.~\ref{eq:original_conditional} as:
\begin{align}
\label{eq:rewritten_original_conditional}
   p_{\theta}(x_{i},a_{i}|&z_{x},z_{a})=p_{\theta_{x}}(x_{i}|z_{x},z_{a})p_{\theta_{a}}(a_{i}|z_{x},z_{a}) \nonumber \\
    &=p_{\theta_{x}}(x_{i}|z_{x})p_{\theta_{x}}(x_{i}|z_{a})p_{\theta_{a}}(a_{i}|z_{a})p_{\theta_{a}}(a_{i}|z_{x})
\end{align}
where $\theta_{x}$ is the parameter of the shared decoder $D_{X}$ and $\theta_{a}$ is the parameter of the shared decoder $D_{A}$. Eq.~\ref{eq:rewritten_original_conditional} indicates a paired structure-attribute matching strategy where $z_{x}$ is used to reconstruct $x_{i}$ and $a_{i}$, and $z_{a}$ is used to reconstruct $a_{i}$ and $x_{i}$. 

Taking the above into consideration, the joint reconstruction loss in the paired structure-attribute matching strategy is written as:
\begin{align}
\label{eq:objective_Lr}
 \min_{\theta_{x},\theta_{a},\phi_{x},\phi_{a}} \mathcal{L}_{r} = &-\mathbb{E}_{x_{i}\sim p_{X}}[\mathbb{E}_{q_{\phi_{x}}(z_{x}|x_{i})}[\log p_{\theta_{x}}(x_{i}|z_{x})]] \nonumber \\ 
 &-\mathbb{E}_{a_{i}\sim p_{A}}[\mathbb{E}_{q_{\phi_{a}}(z_{a}|a_{i})}[\log p_{\theta_{a}}(a_{i}|z_{a})]]  \nonumber\\
 &-\lambda_\mathrm{c} \cdot \mathbb{E}_{a_{i}\sim p_{A}}[\mathbb{E}_{q_{\phi_{a}}(z_{a}|a_{i})}[\log p_{\theta_{x}}(x_{i}|z_{a})]] \nonumber \\ 
 &-\lambda_\mathrm{c}\cdot \mathbb{E}_{x_{i}\sim p_{X}} [\mathbb{E}_{q_{\phi_{x}}(z_{x}|x_{i})}[\log p_{\theta_{a}}(a_{i}|z_{x})]].
\end{align}
The first two terms in Eq.\ref{eq:objective_Lr} represent the self-reconstruction stream. It means the information from attributes (\textit{resp.} structures) is decoded as attributes (\textit{resp.} structures).
The last two terms in Eq.\ref{eq:objective_Lr} indicate the cross-reconstruction stream. It means the information from attributes (\textit{resp.} structures) is decoded as attributes (\textit{resp.} structures). $\lambda_{c}$ is a hyper-parameter to weight the cross reconstruction stream.

\subsubsection{Adversarial Distribution Matching}
Term $D_{KL}[q_{\phi}(z_{x},z_{a}|x_{i},a_{i})||p(z_{x},z_{a})]$ in Eq.~\ref{eq:ELBO} indicates a prior distribution matching for the latent factors $z_{x}$ and $z_{a}$. Learning joint prior is a specific topic that could be explored in the future with the guidance of recent works~\cite{tomczak2017vae,rezende2015variational,yin2018semi}, here we set $p(z_{x},z_{y})=p(z)p(z)$ for simplicity, where $p(z)$ is the prior distribution for the \textit{shared-latent space}.
Taking Eq.~\ref{eq:rewritten_q_phi} into consideration, the prior regularization term is formulated as:
\begin{align}
\label{eq:kl_new_form}
 D_{KL}[q_{\phi}(z_{x},z_{a}|x_{i},a_{i})||p(z_{x},z_{a})]&=D_{KL}[q_{\phi_{x}}(z_{x}|x_{i})||p(z)]\nonumber \\
    +D_{KL}&[q_{\phi_{a}}(z_{a}|a_{i})||p(z)]
\end{align}
Eq.~\ref{eq:kl_new_form} states the regularization that matches $q_{\phi_{x}}(z_{x}|x_{i})$ and $q_{\phi_{a}}(z_{a}|a_{i})$ to prior $p(z)$.

Since it is not easy to derive explicit formulations for some complex priors in KL-divergence, SAT employs the adversarial distribution matching that can impose an arbitrary prior distribution for the latent codes without hard derivation~\cite{makhzani2015adversarial}. Following~\cite{makhzani2015adversarial}, the prior regularization loss in adversarial distribution matching is written as:
\begin{align}
\label{eq:adversarial_loss}
  \min_{\psi} \max_{\phi_{x},\phi_{a}} \mathcal{L}_{adv}=&- \mathbb{E}_{z_{p}\sim p(z)}[\log \mathcal{D}(z_{p})] \nonumber\\
  &-\mathbb{E}_{z_{x}\sim q_{\phi_{x}}(z_{x}|x_{i})}[\log (1-\mathcal{D}(z_{x}))] \nonumber \\
  &-\mathbb{E}_{z_{p}\sim p(z)}[\log \mathcal{D}(z_{p})] \nonumber\\
  &-\mathbb{E}_{z_{a}\sim q_{\phi_{a}}(z_{a}|a_{i})}[\log (1-\mathcal{D}(z_{a}))]
\end{align}
where $\psi$ is the parameters of the shared discriminator $\mathcal{D}$. $z_{p}$ indicates true samples sampled from the prior $p(z)$ for adversarial learning.

The employed adversarial distribution matching in Eq.~\ref{eq:adversarial_loss} has several advantages compared to the KL-divergence in Eq.~\ref{eq:kl_new_form}. The KL-divergence tries to match $q_{\phi_{x}}(z_{x}|x_{i})$ to prior $p(z)$, which will have risk to lose the information from input $x_{i}$. By contrast,
the adversarial distribution matching in latent space makes the posterior $q_{\phi_{x}}(z_{x}|x_{i})$ to be the aggregated posterior $q_{\phi_{x}}(z_{x})$, which encourages $z_{x}$ to match the whole distribution of $p(z)$~\cite{makhzani2015adversarial,makhzani2018implicit}. Accordingly, $z_{a}$ can match the whole distribution of $p(z)$ in similar way. Meanwhile, the mode collapse problem in adversarial learning could be avoided since our method involves a reconstruction operation which encourages the latent embeddings to match both the prior and the entire true data distribution~\cite{srivastava2017veegan}. 

\subsubsection{Objective Function and Implementation} 
Taking the above into summary, the objective function of SAT is:
\begin{align}
\label{eq:objective}
 \min_{\Theta}\max_{\Phi} \mathcal{L} =\mathcal{L}_{r}+\mathcal{L}_{{adv}}
\end{align} 
where $\Theta=\{\theta_{x},\theta_{a},\phi_{x},\phi_{a},\psi\}$ and $\Phi=\{\phi_{x},\phi_{a}\}$. By leveraging the~\textit{shared-latent space} assumption, SAT leverages the correlation between observed node attributes and structures and thus facilitates learning on attribute-missing graphs. It is worthwhile to point out that SAT cannot handle attribute-incomplete graphs. On attribute-incomplete graphs, the incomplete node attributes would cause too much information gap for structures and attributes, and SAT based on the \emph{shared-latent space} assumption is not appropriate.

For the implementation, SAT consists of three components: (1) self-reconstruction stream, (2) cross-reconstruction stream, and (3) adversarial distribution matching. In the self-reconstruction stream, both the observed attributes $X^{o}$ and structures $A$ are encoded as latent codes $Z_{X^{o}}$ and $Z_{A}$, respectively. Then $Z_{X^{o}}$ and $Z_{A}$ are decoded as $X^{o}$ and $A$, respectively. In the cross-reconstruction stream, the structure embedding $Z_{A^{o}}$ of attribute-observed nodes are obtained through an embedding lookup layer from $Z_{A}$.
Then, the latent codes $Z_{X^{o}}$ and $Z_{A^{o}}$ are decoded as $A^{o}$ and $X^{o}$, respectively. These appear in Figure~\ref{figure:model_architecture}, in which $E_{X}$ is a two-layer MLP, $E_{A}$ is a two-layer GNN, $D_{X}$ is a two-layer MLP and $D_{A}$ is a two-layer MLP followed by a \textit{sigmoid} function. In the adversarial distribution matching module, we apply adversarial learning between $Z_{X^{o}}$, $Z_{A}$ and samples from a standard normal distribution $\mathcal{N}(0,1)$, sharing the same two-layer MLP discriminator. And \textit{Relu} is used as the non-linear activation function for all three modules.

\subsection{Time Complexity Analysis}
Stochastic training of DNN methods involves two steps, the forward and backward computations.
SAT includes a paired structure-attribute matching strategy and an adversarial distribution matching strategy. We thus decompose the time complexity of SAT into two parts, namely the time complexity of the paired structure-attribute matching and the time complexity of the adversarial distribution matching. 

In the paired structure-attribute matching, a GNN backbone encodes the structures into latent codes and a MLP module encodes the attributes into latent embeddings. Following the analysis of GNN models in~\cite{wu2019comprehensive}, the time complexity of the GNN backbone is $\mathcal{O}(|\mathcal{E}|)$, where $|\mathcal{E}|$ denote the number of edges. And for the MLP module that encodes node attributes, the time complexity is $\mathcal{O}(N_{o}F)$, where $N_{o}$ is the number of attribute-observed nodes and $F$ is the attribute dimension. Since the GNN backbone and the MLP module can be calculated parallelly, then the time complexity for the paired structure-attribute matching is $\mathcal{O}(\max\{|\mathcal{E}|, N_{o}F\})$.
After the calculation of the paired structure-attribute matching, the adversarial distribution matching can be conducted to impose a prior distribution on the latent codes. The time complexity for the adversarial distribution matching is $\mathcal{O}(N_{o}d)$, where $d$ is the latent dimension.

In summary, the time complexity of the non-parallel SAT is $\mathcal{O}(|\mathcal{E}|+N_{o}F+N_{o}d)$, and the parallel counterpart is $\mathcal{O}(\max\{|\mathcal{E}|, N_{o}F\}+N_{o}d)$. Note that SAT has a flexible GNN backbone such as GCN~\cite{kipf2016semi} and GraphSage~\cite{hamilton2017inductive}.
This indicates SAT can be scalable to much larger datasets when a scalable GNN backbone is applied.

\begin{table*}[]
\caption{The statistics of seven datasets. In this table, ``attribute form'' means the attribute style.     ``$\#$avg hot num'' means the average hot number for multi-hot attributes of nodes. \#class indicates the number of categories.}
\centering
\label{table:dataset}
\renewcommand{\arraystretch}{1.0}
 \setlength{\tabcolsep}{1.0mm}{ 
  \scalebox{1.2}{
\begin{tabular}{c|ccccccc}
\hline
                & Cora        & Citeseer    & Steam       & Pubmed      & \begin{tabular}[c]{@{}c@{}}Coauthor\\ -CS\end{tabular} & \begin{tabular}[c]{@{}c@{}}Amazon\\ -Computer\end{tabular} & \begin{tabular}[c]{@{}c@{}}Amazon\\ -Photo\end{tabular} \\ \hline
\#nodes         & 2,708       & 3,327       & 9,944       & 19,717      & 18,333                                                 & 13,752                                                     & 7,650                                                   \\
\#edges         & 5,278       & 4,228       & 266,981     & 44,324      & 81,894                                                 & 245,861                                                    & 119,081                                                 \\
\#graph density & 0.07\%      & 0.04\%      & 0.26\%      & 0.01\%      & 0.02\%                                                 & 0.13\%                                                     & 0.20\%                                                  \\
\#attribute dim & 1,433       & 3,703       & 352         & 500         & 6,805                                                   & 767                                                        & 745                                                     \\
\#avg hot num   & 18.17       & 31.6        & 8.45        & -           & -                                                      & 267.23                                                     & 258.81                                                  \\
\#class         & 7           & 6           & -           & 3           & 15                                                     & 10                                                         & 8                                                       \\
attribute form  & categorical & categorical & categorical & real-valued & real-valued                                            & categorical                                                & categorical                                             \\ \hline
\end{tabular}
}}
\end{table*}

\section{Experiments}
In this section, we will first introduce the datasets and baselines as well as the experimental settings in our paper. Then, the experiments for \emph{node attribute completion} and link prediction are introduced. We further provide some visualization results for better comprehension of SAT.  
\subsection{Datasets} 
We evaluate the proposed SAT on seven real-world datasets to quantify the performance.
\begin{itemize}
 \item Cora. Cora~\cite{mccallum2000automating} is a citation graph with 2,708 papers as nodes and 10,556 citation links as edges. The attribute vector of each node indicates whether the corresponding paper contains specific word tokens, and it is represented as a multi-hot vector with dimension 1,433.
 \item Citeseer. Citeseer~\cite{sen2008collective} is also a citation graph which contains 9,228 edges and 3,327 papers. After processing of the content, 3,703 distinct words compose the attribute corpus. Each attribute vector is formed from the corpus and expressed by a multi-hot vector with dimension 3,703.
 \item Steam. Steam is a dataset collected from a game website with user-bought behaviors of 9,944 items and 352 tags. We count the co-purchase frequency between every two games and make a sparse item co-purchase graph through binarization operation with the threshold as 10. After that, we obtain 533,962 edges for this graph. The tag corpus constructs the multi-hot attribute vector for each item with dimension 352.
 \item Pubmed. Pubmed~\cite{namata2012query} is a citation graph with 19,717 nodes and 88,651 edges. Each attribute vector is described by a Term Frequency-Inverse Document Frequency (TF-IDF) vector from 500 distinct terms.
 \item Coauthor-CS. Coauthor-CS~\cite{shchur2018pitfalls} is a coauthor dataset based on the Microsoft Academic Graph from the KDD Cup 2016 challenge. In this dataset, we have 18,333 nodes that are authors and 81,894 edges that mean two authors are connected if they coauthored a paper. Node attributes are frequencies of the 6,805 keywords from each author's papers. 
 \item Amazon-Computer and Amazon-Photo. Amazon-Computer~\cite{shchur2018pitfalls} and Amazon-Photo~\cite{shchur2018pitfalls} are from segments of the Amazon co-purchase graph~\cite{mcauley2015image}. In Amazon-computer, there are 13,752 items and 245,861 edges. In Amazon-Photo, there are 7,650 items and 119,081 edges. For both datasets, multi-hot bag-of-words encoded from the product reviews construct the multi-hot node attributes.
\end{itemize}
Among these datasets, attributes of Cora, Citeseer, Steam, Amazon-Computer and Amazon-Photo are categorical and represented as multi-hot vectors. For Pubmed and Coauthor-CS, attributes are real-valued and represented as scalars. More details are shown in Table~\ref{table:dataset}.

\subsection{Baselines} 
\textbf{Node Attribute Completion.}
Since there is no specialized method for the \emph{node attribute completion} task, we compare our SAT with the following representative baselines selected from five aspects: NeighAggre is from the~\emph{classical aggregation} aspect, VAE is from the~\emph{auto-encoding} aspect, GNN methods are from the~\emph{GNN-encoding} aspect, GraphRNA and ARWMF are from the~\emph{attributed random walking} aspect and Hers is from the~\emph{cold-start recommendation} aspect.
\begin{itemize}
 \item NeighAggre. NeighAggre~\cite{csimcsek2008navigating} aggregates the neighbors' attributes for nodes without attributes through mean pooling. NeighAggre is a classical profiling algorithm, which is simple, but usually provides strong empirical performance. When a neighbor' attributes are missing, we do not regard it as the aggregation node. We use the one-hop neighbors as a node's neighbors here. 
 \item VAE. VAE~\cite{kingma2013auto} is one well-known generative method. In our setting, we make normal VAE for the attribute-observed nodes, encoding attributes of these nodes as latent codes. For test nodes without any attribute, we use neighbor aggregation like NeighAggre in the latent space to obtain the aggregated latent codes for test nodes. Then, the decoder in VAE can be used to restore attributes for test nodes.
 \item GCN, GraphSage and GAT. GCN~\cite{kipf2016semi}, GraphSage~\cite{hamilton2017inductive} and GAT~\cite{velivckovic2017graph} are three representative GNN methods in recent years. In our problem, only graph structures are used as input and encoded as latent codes. Then the latent codes are decoded as node attributes. In the test stage, the latent codes of test nodes are used to restore attributes by the decoder of GNN. Note that we do not use the zero-filling trick here since this task require reconstruction of attributes and the zero-filling trick largely deteriorates the performance. 
 \item GraphRNA~\cite{huang2019graph} and ARWMF~\cite{chen2019} are two representative attributed random walk based methods. In attributed random walk based methods, graphs with node attributes are taken as bipartite graphs and then random walks based approaches are applied to learn node embeddings. Thereby, attributed random walk based methods can potentially work on attribute-missing graphs. Note that there are three variants of ARWMF in~\cite{chen2019}, we introduce the third variant as the baseline since it generally shows better performance than the other two variants~\cite{chen2019}.
 \item Hers. Since the attribute-completion on attribute-missing graphs is similar to the cold-start recommendation problem. We thus introduce one representative cold-start recommendation method called Hers~\cite{hu2019hers} to be a comparison method.
\end{itemize}

\textbf{Link Prediction.}
Even on attribute-missing graphs, link prediction is still an important task. We choose baselines for link prediction comparison from two aspects: the \emph{structure-only} aspect and the \emph{fused-structure-attribute} aspect. SPM, DeepWalk, Node2Vec and Hers are only suitable for the \emph{structure-only} aspect. GraphRNA and ARWMF are only feasible for the \emph{fused-structure-attribute} aspect. GAE and VGAE are feasible for both aspects.
\begin{itemize}
 \item SPM. Structure perturbation method (SPM)~\cite{lu2015toward} is one classic link prediction model on graphs. It assumes that the regularity of a graph is reflected in the consistency of structures before and after structural perturbations. Based on this assumption, it proposes a universal structural consistency index for link prediction on graphs.
 \item DeepWalk. DeepWalk~\cite{perozzi2014deepwalk} learns node representation from random walk sequences on graphs. It only considers graph structures in the learning process.
 \item Node2vec. Node2vec~\cite{grover2016node2vec} extends DeepWalk by designing a biased random walk to control the Bread First Search (BFS) and Deep First Search (DFS). It also learns node embeddings with only graph structures.
 \item GAE. Graph Auto-encoder (GAE)~\cite{kipf2016variational} combines GCN and the auto-encoder theory. Specifically, GCN plays as a graph convolution encoder to replace the MLP encoder in vanilla auto-encoder~\cite{kingma2013auto}. When GAE works for the \emph{fused-structure-attribute} aspect, it requires structures and attributes as a whole input. Thus we use zero-filling, a common trick in data mining community, for the attribute-missing nodes. Furthermore, we also use GAT as the graph convolution encoder called GAE(GAT) as another baseline method.
 \item VGAE. Variational Graph Auto-encoder (VGAE)~\cite{kipf2016variational} performs variational inference for GAE with reparameterization tricks. We use it in the same way as GAE and also combine it with GAT called VGAE(GAT) as another baseline.
 \item Hers, GraphRNA and ARWMF. Since Hers~\cite{hu2019hers}, GraphRNA~\cite{huang2019graph} and ARWMF~\cite{chen2019} are able to learn node embeddings, we thus also introduce them here for the link prediction task.
\end{itemize}

\subsection{Experimental Setup}
For \emph{node attribute completion}, we randomly sample $40\%$ nodes with attributes as training data and $10\%$ as validation data and the rest $50\%$ as test data. For link prediction, we randomly sample $60\%$, $20\%$ and $20\%$ links with equivalent non-links as train, validation and test data, respectively. 

For the cold-start recommendation method Hers~\cite{hu2019hers}, we use the graph data as the social context and the attribute matrix as the target to complete.
For random walk based methods (DeepWalk\footnote{https://github.com/phanein/deepwalk}, Node2Vec\footnote{https://github.com/aditya-grover/node2vec}) and attributed random walk based methods (GraphRNA and ARWMF), their hyper-parameters such as number of walks, walk length and window size are set as the default according to the codes online. All GNN methods use a two-layer graph convolution. For GraphSage, we set the neighborhood sampling size is (5,5) for Cora, Citeseer and Steam, while (10,25) for Pubmed according to the default settings of the online codes\footnote{https://github.com/williamleif/graphsage-simple/}. For GraphSage on Coauthor-CS, Amazon-Computer and Amazon-Photo, we found it performs well when neighborhood sampling size is (10,25).
For GAT\footnote{https://github.com/Diego999/pyGAT}, in order to reduce training time and storage, we use one-head attention.  
By following recent works~\cite{kipf2016semi,velivckovic2017graph}, we set the latent dimension as 64 for all learning-based methods. Learning rate is 0.005. Dropout rate equals 0.5, and the maximum iteration number is 1,000. Adam optimizer is applied for them to learn the model parameters.  

For our distribution matching based SAT, we set the generative step as 2 and discriminative step as 1. For balanced comparison, the hyper-parameters of different GNN backbones are the same as the GNN baselines.
The hyper-parameter $\lambda_\mathrm{c}$ ranges in $[0.1,1.0,2.0,5.0,10.0,20.0,50.0,100.0]$. According to the performance on the validation set, for the \emph{node attribute completion} task, we have $\lambda_\mathrm{c}=10.0, 10.0, 50.0, 10.0, 100.0, 100.0, 100.0$ for Cora, Citeseer, Pubmed, Steam, Coauthor-CS, Amazon-Computer and Amazon-Photo, respectively. For the link prediction task, we set $\lambda_\mathrm{c}=10.0, 10.0, 1.0, 10.0, 0.1, 0.1, 0.1$ for Cora, Citeseer, Pubmed, Steam, Coauthor-CS, Amazon-Computer and Amazon-Photo, respectively.
For all methods, The best trained model is chosen for testing according to the performance on the validation set. 

For datasets with categorical attributes, weighted Binary Cross Entropy loss (BCE) is applied. The weight put on non-zero values equals $\frac{\#zero~count}{\#non-zero~count}$ which is calculated from the training attribute matrix. For datasets with real-valued attributes, Mean Square Error (MSE) loss is used. We do not use GraphSage as the baseline for link prediction since it takes much time for random neighborhood sampling in each iteration. The experiments are conducted in 5 times, and then the mean value is adopted as the performance. The method is implemented by Pytorch on a machine with one Nvidia TitanX GPU.

\begin{table*}[ht]
\centering
\caption{Node classification of the \emph{node-level} evaluation for \emph{node attribute completion}. The first column with ``X'', ``A'' and ``A+X'' indicates three settings to do node classification with only attributes, only structures and the fused one. 
Note that SAT has an extendable GNN backbone, so we combine it with different models as SAT(GCN), SAT(GraphSage) and SAT(GAT).
SAT(GCN)-no self, SAT(GCN)-no cross and SAT(GCN)-no adver respectively denotes SAT without self-reconstruction terms, cross-reconstruction terms and adversarial learning terms. The term ``True attributes'' indicates we use the ground truth attributes to do node classification.}
\label{table:classification}
\vspace{0pt}
\renewcommand{\arraystretch}{1.0}
 \setlength{\tabcolsep}{1.5mm}{ 
  \scalebox{1.0}{
\begin{tabular}{c|cccccccc}
\hline
                       & \begin{tabular}[c]{@{}c@{}}attribute \\ completion method\end{tabular} & \begin{tabular}[c]{@{}c@{}}classification \\ method\end{tabular} & Cora            & Citeseer        & Pubmed                                 & \begin{tabular}[c]{@{}c@{}}Coauthor\\ -CS\end{tabular} & \begin{tabular}[c]{@{}c@{}}Amazon\\ -Computer\end{tabular} & \begin{tabular}[c]{@{}c@{}}Amazon\\ -Photo\end{tabular} \\ \hline
                       & NeighAggre                                                             & MLP                                                              & 0.6248          & 0.5539          & 0.5150                                 & 0.7562                                                 & 0.8365                                                     & 0.8846                                                  \\
                       & VAE                                                                    & MLP                                                              & 0.2826          & 0.2551          & 0.4008                                 & 0.2317                                                 & 0.3747                                                     & 0.2598                                                  \\
                       & GCN                                                                    & MLP                                                              & 0.3943          & 0.3768          & 0.3992                                 & 0.2180                                                 & 0.3660                                                     & 0.2683                                                  \\
                       & GraphSage                                                              & MLP                                                              & 0.4852          & 0.3933          & 0.4013                                 & 0.2317                                                 & 0.3747                                                     & 0.2598                                                  \\
                       & GAT                                                                    & MLP                                                              & 0.4143          & 0.2129          & 0.3996                                 & 0.2317                                                 & 0.3747                                                     & 0.2598                                                  \\
                       & Hers                                                                   & MLP                                                              & 0.3046          & 0.2585          & 0.4004                                 & 0.2317                                                 & 0.3747                                                     & 0.2598                                                  \\
                       & GraphRNA                                                               & MLP                                                              & 0.7581          & 0.6320          & 0.6035                                 & {\color[HTML]{333333} \textbf{0.7710}}                 & 0.6968                                                     & 0.8407                                                  \\
                       & ARWMF                                                                  & MLP                                                              & 0.7769          & 0.2267          & {\color[HTML]{000000} \textbf{0.6180}}  & 0.2320                                                 & 0.5608                                                     & 0.4675                                                  \\ \cline{2-9} 
                       & SAT(GCN)-no self                                                       & MLP                                                              & 0.7074          & 0.4976          & 0.4000                                 & 0.7504                                                 & 0.7410                                                     & 0.8585                                                  \\
                       & SAT(GCN)-no cross                                                      & MLP                                                              & 0.3036          & 0.2289          & 0.4023                                 & 0.2317                                                 & 0.3748                                                     & 0.2613                                                  \\
                       & SAT(GCN)-no adver                                                      & MLP                                                              & 0.7587          & 0.6051          & 0.4680                                 & 0.6879                                                 & 0.7356                                                     & 0.8629                                                  \\
                       & SAT(GCN)                                                               & MLP                                                              & 0.7644          & 0.6010          & 0.4652                                 & 0.7592                                                 & {\color[HTML]{333333} 0.7410}                              & {\color[HTML]{333333} 0.8762}                           \\
                       & SAT(GraphSage)                                                         & MLP                                                              & 0.7032          & 0.5936          & 0.4585                                 & 0.6637                                                 & {\color[HTML]{333333} \textbf{0.8396}}                     & {\color[HTML]{333333} \textbf{0.9035}}                  \\
                       & SAT(GAT)                                                               & MLP                                                              & \textbf{0.7937} & \textbf{0.6475} & 0.4618                                 & 0.7672                                                 & 0.8201                                                     & 0.8976                                                  \\ \cline{2-9} 
\multirow{-15}{*}{X}   & True attributes                                                        & MLP                                                              & 0.7618          & 0.7174          & 0.656                                  & 0.9396                                                 & 0.8423                                                     & 0.9151                                                  \\ \hline
                       & -                                                                      & DeepWalk+MLP                                                     & 0.7149          & 0.4802          & 0.6917                                 & 0.7561                                                 & 0.8444                                                     & 0.8955                                                  \\
                       & -                                                                      & Node2Vec+MLP                                                     & 0.6830          & 0.4422          & 0.6721                                 & 0.7554                                                 & 0.8415                                                     & 0.8908                                                  \\
\multirow{-3}{*}{A}    & -                                                                      & GCN                                                              & \textbf{0.7631} & \textbf{0.5651} & \textbf{0.7125}                        & {\color[HTML]{333333} \textbf{0.8370}}                 & {\color[HTML]{333333} \textbf{0.8785}}                     & {\color[HTML]{333333} \textbf{0.9117}}                  \\ \hline
                       & NeighAggre                                                             & GCN                                                              & 0.6494          & 0.5413          & 0.6564                                 & 0.8031                                                 & 0.8715                                                     & 0.901                                                   \\
                       & VAE                                                                    & GCN                                                              & 0.3011          & 0.2663          & 0.4007                                 & 0.2335                                                 & 0.4023                                                     & 0.3781                                                  \\
                       & GCN                                                                    & GCN                                                              & 0.4387          & 0.4079          & 0.4203                                 & 0.2180                                                 & 0.3974                                                     & 0.3656                                                  \\
                       & GraphSage                                                              & GCN                                                              & 0.5779          & 0.4278          & 0.4200                                 & 0.2335                                                 & 0.4019                                                     & 0.3784                                                  \\
                       & GAT                                                                    & GCN                                                              & 0.4525          & 0.2688          & 0.4196                                 & 0.2334                                                 & 0.4034                                                     & 0.3789                                                  \\
                       & Hers                                                                   & GCN                                                              & 0.3405          & 0.3229          & 0.4205                                 & 0.2334                                                 & 0.4025                                                     & 0.3794                                                  \\
                       & GraphRNA                                                               & GCN                                                              & 0.8198          & 0.6394          & {\color[HTML]{333333} \textbf{0.8172}} & {\color[HTML]{333333} \textbf{0.8851}}                 & 0.8650                                                     & 0.9207                                                  \\
                       & ARWMF                                                                  & GCN                                                              & 0.8025          & 0.2764          & 0.8089                                 & 0.8347                                                 & 0.7400                                                     & 0.6146                                                  \\ \cline{2-9} 
                       & SAT(GCN)-no self                                                       & GCN                                                              & 0.7727          & 0.5358          & 0.4197                                 & 0.8575                                                 & 0.8455                                                     & 0.9127                                                  \\
                       & SAT(GCN)-no cross                                                      & GCN                                                              & 0.3402          & 0.2698          & 0.4204                                 & 0.3499                                                 & 0.4394                                                     & 0.3846                                                  \\
                       & SAT(GCN)-no adver                                                      & GCN                                                              & 0.8231          & 0.6609          & 0.7371                                 & 0.8161                                                 & 0.8439                                                     & 0.9123                                                  \\
                       & SAT(GCN)                                                               & GCN                                                              & 0.8327          & 0.6599          & 0.7537                                 & 0.8576                                                 & 0.8519                                                     & 0.9163                                                  \\
                       & SAT(GraphSage)                                                         & GCN                                                              & 0.8255          & 0.6547          & 0.7360                                 & 0.8001                                                 & {\color[HTML]{333333} \textbf{0.8834}}                     & 0.9234                                                  \\
                       & SAT(GAT)                                                               & GCN                                                              & \textbf{0.8579} & \textbf{0.6767} & 0.7439                                 & 0.8402                                                 & 0.8766                                                     & {\color[HTML]{333333} \textbf{0.9260}}                  \\ \cline{2-9} 
\multirow{-15}{*}{A+X} & True attributes                                                        & GCN                                                              & 0.8493          & 0.7348          & 0.8723                                 & 0.9186                                                 & 0.9097                                                     & 0.9412                                                  \\ \hline
\end{tabular}
}}
\end{table*}

\subsection{Node Attribute Completion}
\subsubsection{Necessity of \emph{Node Attribute Completion}}
\emph{Node attribute completion} is a new problem on graphs and distinguishes from other existing problems including label prediction and graph embedding. 

Although node labels could be one kind of attributes, \emph{node attribute completion} is quite different from the widely explored label prediction problem on graphs~\cite{kipf2016semi} due to their \textit{characteristic}, \textit{functionality} and \textit{methodology}. From the aspect of \textit{characteristic}, labels are designed to be clean and only limited to categorical values, while attributes are more general, noisy and either numeric or categorical. From the aspect of \textit{functionality}, label prediction on graphs is one subsequent task of \emph{node attribute completion}.
Many label prediction algorithms require both structures and attributes as inputs to achieve superior performance~\cite{zhang2018deep}. \emph{Node attribute completion} enables recovering unknown attributes of nodes, which is useful for many subsequent tasks, including label prediction. From the aspect of \textit{methodology}, when typical label-prediction methods are applied to \emph{node attribute completion}, such as GCN~\cite{kipf2016semi} used to predict labels (categorical attributes) by minimizing classiﬁcation loss, they ignore the joint relationship between node structures and attributes. Furthermore, compare to the graph embedding problem~\cite{perozzi2014deepwalk,grover2016node2vec,kipf2016variational}, \emph{node attribute completion} restores high-dimensional and human-understandable attributes while graph embedding learns to represent nodes or graphs in low-dimensional and hard-perceptive vectors.

\begin{table*}[!ht]
\small
\centering
\caption{Profiling of the \emph{attribute-level} evaluation for \emph{node attribute completion}. Note that we use top 3, 5, 10 to evaluate the performance on Steam since the average non-zero hot number of node attributes on Steam is quite small.}
\label{table:label_propagation}
\vspace{0pt}
\renewcommand{\arraystretch}{1.0}
 \setlength{\tabcolsep}{1.5mm}{ 
  \scalebox{0.9}{
\begin{tabular}{c|ccccccc}
\hline
\multicolumn{1}{l|}{}              & \multicolumn{1}{c|}{Method}            & Recall@10                     & Recall@20                     & Recall@50                     & NDCG@10                       & NDCG@20                       & NDCG@50                       \\ \hline
                                   & \multicolumn{1}{c|}{NeighAggre}        & 0.0906                        & 0.1413                        & 0.1961                        & 0.1217                        & 0.1548                        & 0.1850                        \\
                                   & \multicolumn{1}{c|}{VAE}               & 0.0887                        & 0.1228                        & 0.2116                        & 0.1224                        & 0.1452                        & 0.1924                        \\
                                   & \multicolumn{1}{c|}{GCN}               & 0.1271                        & 0.1772                        & 0.2962                        & 0.1736                        & 0.2076                        & 0.2702                        \\
                                   & \multicolumn{1}{c|}{GraphSage}         & 0.1284                        & 0.1784                        & 0.2972                        & 0.1768                        & 0.2102                        & 0.2728                        \\
                                   & \multicolumn{1}{c|}{GAT}               & 0.1350                        & 0.1812                        & 0.2972                        & 0.1791                        & 0.2099                        & 0.2711                        \\
                                   & \multicolumn{1}{c|}{Hers}              & 0.1226                        & 0.1723                        & 0.2799                        & 0.1694                        & 0.2031                        & 0.2596                        \\
                                   & \multicolumn{1}{c|}{GraphRNA}          & 0.1395                        & 0.2043                        & 0.3142                        & 0.1934                        & 0.2362                        & 0.2938                        \\
                                   & \multicolumn{1}{c|}{ARWMF}             & 0.1291                        & 0.1813                        & 0.296                         & 0.1824                        & 0.2182                        & 0.2776                        \\ \cline{2-8} 
                                   & \multicolumn{1}{c|}{SAT(GCN)-no self}  & 0.1378                        & 0.2018                        & 0.3339                        & 0.1931                        & 0.2360                        & 0.3052                        \\
                                   & \multicolumn{1}{c|}{SAT(GCN)-no cross} & 0.1224                        & 0.1724                        & 0.2823                        & 0.1686                        & 0.2023                        & 0.2599                        \\
                                   & \multicolumn{1}{c|}{SAT(GCN)-no adver} & 0.1356                        & 0.1966                        & 0.3179                        & 0.1924                        & 0.2331                        & 0.2971                        \\
                                   & \multicolumn{1}{c|}{SAT(GCN)}          & 0.1508                        & 0.2182                        & 0.3429                        & 0.2112                        & 0.2546                        & 0.3212                        \\
                                   & \multicolumn{1}{c|}{SAT(GraphSage)}    & 0.1356                        & 0.1981                        & 0.3165                        & 0.1905                        & 0.2320                        & 0.2947                        \\
\multirow{-14}{*}{Cora}            & \multicolumn{1}{c|}{SAT(GAT)}          & \textbf{0.1653}               & \textbf{0.2345}               & \textbf{0.3612}               & \textbf{0.2250}               & \textbf{0.2723}               & \textbf{0.3394}               \\ \hline
                                   & \multicolumn{1}{c|}{NeighAggre}        & 0.0511                        & 0.0908                        & 0.1501                        & 0.0823                        & 0.1155                        & 0.1560                        \\
                                   & \multicolumn{1}{c|}{VAE}               & 0.0382                        & 0.0668                        & 0.1296                        & 0.0601                        & 0.0839                        & 0.1251                        \\
                                   & \multicolumn{1}{c|}{GCN}               & 0.0620                        & 0.1097                        & 0.2052                        & 0.1026                        & 0.1423                        & 0.2049                        \\
                                   & \multicolumn{1}{c|}{GraphSage}         & 0.0612                        & 0.1097                        & 0.2058                        & 0.1003                        & 0.1393                        & 0.2034                        \\
                                   & \multicolumn{1}{c|}{GAT}               & 0.0561                        & 0.1012                        & 0.1957                        & 0.0878                        & 0.1253                        & 0.1872                        \\
                                   & \multicolumn{1}{c|}{Hers}              & 0.0576                        & 0.1025                        & 0.1973                        & 0.0904                        & 0.1279                        & 0.1900                        \\
                                   & \multicolumn{1}{c|}{GraphRNA}          & 0.0777                        & 0.1272                        & 0.2271                        & 0.1291                        & 0.1703                        & 0.2358                        \\
                                   & \multicolumn{1}{c|}{ARWMF}             & 0.0552                        & 0.1015                        & 0.1952                        & 0.0859                        & 0.1245                        & 0.1858                        \\ \cline{2-8} 
                                   & \multicolumn{1}{c|}{SAT(GCN)-no self}  & 0.0679                        & 0.1163                        & 0.2140                        & 0.1167                        & 0.1570                        & 0.2209                        \\
                                   & \multicolumn{1}{c|}{SAT(GCN)-no cross} & 0.0564                        & 0.1013                        & 0.1963                        & 0.0863                        & 0.1238                        & 0.1860                        \\
                                   & \multicolumn{1}{c|}{SAT(GCN)-no adver} & 0.0705                        & 0.1168                        & 0.2145                        & 0.1196                        & 0.1582                        & 0.2223                        \\
                                   & \multicolumn{1}{c|}{SAT(GCN)}          & 0.0764                        & 0.1280                        & 0.2377                        & 0.1298                        & 0.1729                        & 0.2447                        \\
                                   & \multicolumn{1}{c|}{SAT(GraphSage)}    & 0.0704                        & 0.1163                        & 0.2174                        & 0.1179                        & 0.1563                        & 0.2227                        \\
\multirow{-14}{*}{Citeseer}        & \multicolumn{1}{c|}{SAT(GAT)}          & \textbf{0.0811}               & \textbf{0.1349}               & \textbf{0.2431}               & \textbf{0.1385}               & \textbf{0.1834}               & \textbf{0.2545}               \\ \hline
                                   & \multicolumn{1}{c|}{NeighAggre}        & 0.0321                        & 0.0593                        & 0.1306                        & 0.0788                        & 0.1156                        & 0.1923                        \\
                                   & \multicolumn{1}{c|}{VAE}               & 0.0255                        & 0.0502                        & 0.1196                        & 0.0632                        & 0.0970                        & 0.1721                        \\
                                   & \multicolumn{1}{c|}{GCN}               & 0.0273                        & 0.0533                        & 0.1275                        & 0.0671                        & 0.1027                        & 0.1824                        \\
                                   & \multicolumn{1}{c|}{GraphSage}         & 0.0269                        & 0.0528                        & 0.1278                        & 0.0664                        & 0.1020                        & 0.1822                        \\
                                   & \multicolumn{1}{c|}{GAT}               & 0.0271                        & 0.0530                        & 0.1278                        & 0.0673                        & 0.1028                        & 0.1830                        \\
                                   & \multicolumn{1}{c|}{Hers}              & 0.0273                        & 0.0525                        & 0.1273                        & 0.0676                        & 0.1025                        & 0.1825                        \\
                                   & \multicolumn{1}{c|}{GraphRNA}          & 0.0386                        & 0.0690                        & 0.1465                        & 0.0931                        & 0.1333                        & 0.2155                        \\
                                   & \multicolumn{1}{c|}{ARWMF}             & 0.0280                        & 0.0544                        & 0.1289                        & 0.0694                        & 0.1053                        & 0.1851                        \\ \cline{2-8} 
                                   & \multicolumn{1}{c|}{SAT(GCN)-no self}  & 0.039                         & 0.0698                        & 0.1506                        & 0.0956                        & 0.1368                        & 0.2229                        \\
                                   & \multicolumn{1}{c|}{SAT(GCN)-no cross} & 0.0287                        & 0.0554                        & 0.1295                        & 0.0691                        & 0.1054                        & 0.1852                        \\
                                   & \multicolumn{1}{c|}{SAT(GCN)-no adver} & 0.0322                        & 0.0603                        & 0.1364                        & 0.0790                        & 0.1170                        & 0.1984                        \\
                                   & \multicolumn{1}{c|}{SAT(GCN)}          & {\color[HTML]{333333} 0.0391} & {\color[HTML]{333333} 0.0703} & {\color[HTML]{333333} 0.1514} & {\color[HTML]{333333} 0.0963} & {\color[HTML]{333333} 0.1379} & {\color[HTML]{333333} 0.2243} \\
                                   & \multicolumn{1}{c|}{SAT(GraphSage)}    & 0.0419                        & 0.0738                        & 0.1562                        & 0.1030                        & 0.1457                        & 0.2333                        \\
\multirow{-14}{*}{Amazon-Computer} & \multicolumn{1}{c|}{SAT(GAT)}          & \textbf{0.0421}               & \textbf{0.0746}               & \textbf{0.1577}               & \textbf{0.1030}               & \textbf{0.1463}               & \textbf{0.2346}               \\ \hline
                                   & \multicolumn{1}{c|}{NeighAggre}        & 0.0329                        & 0.0616                        & 0.1361                        & 0.0813                        & 0.1196                        & 0.1998                        \\
                                   & \multicolumn{1}{c|}{VAE}               & 0.0276                        & 0.0538                        & 0.1279                        & 0.0675                        & 0.1031                        & 0.1830                        \\
                                   & \multicolumn{1}{c|}{GCN}               & 0.0294                        & 0.0573                        & 0.1324                        & 0.0705                        & 0.1082                        & 0.1893                        \\
                                   & \multicolumn{1}{c|}{GraphSage}         & 0.0295                        & 0.0562                        & 0.1322                        & 0.0712                        & 0.1079                        & 0.1896                        \\
                                   & \multicolumn{1}{c|}{GAT}               & 0.0294                        & 0.0573                        & 0.1324                        & 0.0705                        & 0.1083                        & 0.1892                        \\
                                   & \multicolumn{1}{c|}{Hers}              & 0.0292                        & 0.0574                        & 0.1328                        & 0.0714                        & 0.1094                        & 0.1906                        \\
                                   & \multicolumn{1}{c|}{GraphRNA}          & 0.0390                        & 0.0703                        & 0.1508                        & 0.0959                        & 0.1377                        & 0.2232                        \\
                                   & \multicolumn{1}{c|}{ARWMF}             & 0.0294                        & 0.0568                        & 0.1327                        & 0.0727                        & 0.1098                        & 0.1915                        \\ \cline{2-8} 
                                   & \multicolumn{1}{c|}{SAT(GCN)-no self}  & 0.0399                        & 0.0732                        & 0.1583                        & 0.0982                        & 0.1425                        & 0.2330                        \\
                                   & \multicolumn{1}{c|}{SAT(GCN)-no cross} & 0.0302                        & 0.0578                        & 0.1334                        & 0.0738                        & 0.1112                        & 0.1923                        \\
                                   & \multicolumn{1}{c|}{SAT(GCN)-no adver} & 0.0354                        & 0.0656                        & 0.1463                        & 0.0880                        & 0.1287                        & 0.2149                        \\
                                   & \multicolumn{1}{c|}{SAT(GCN)}          & {\color[HTML]{333333} 0.0410} & {\color[HTML]{333333} 0.0743} & {\color[HTML]{333333} 0.1597} & {\color[HTML]{333333} 0.1006} & {\color[HTML]{333333} 0.145}  & {\color[HTML]{333333} 0.2359} \\
                                   & \multicolumn{1}{c|}{SAT(GraphSage)}    & \textbf{0.0483}               & \textbf{0.0766}               & 0.1601                        & \textbf{0.1082}               & 0.1475                        & 0.2402                        \\
\multirow{-14}{*}{Amazon-Photo}    & \multicolumn{1}{c|}{SAT(GAT)}          & 0.0427                        & 0.0765                        & \textbf{0.1635}               & 0.1047                        & \textbf{0.1498}               & \textbf{0.2421}               \\ \hline
                                   & \multicolumn{1}{c|}{Method}                                 & Recall@3                      & Recall@5                      & Recall@10                     & NDCG@3                        & NDCG@5                        & NDCG@10                       \\ \hline
                                   & \multicolumn{1}{c|}{NeighAggre}                             & 0.0603                        & 0.0881                        & 0.1446                        & 0.0955                        & 0.1204                        & 0.1620                        \\
                                   & \multicolumn{1}{c|}{VAE}                                    & 0.0564                        & 0.0820                        & 0.1251                        & 0.0902                        & 0.1133                        & 0.1437                        \\
                                   & \multicolumn{1}{c|}{GCN}                                    & 0.2392                        & 0.3258                        & 0.4575                        & 0.3366                        & 0.4025                        & 0.4848                        \\
                                   & \multicolumn{1}{c|}{GraphSage}                              & 0.2356                        & 0.3068                        & 0.4568                        & 0.3311                        & 0.3892                        & 0.4817                        \\
                                   & \multicolumn{1}{c|}{GAT}                                    & 0.2395                        & 0.3431                        & 0.4649                        & 0.3364                        & 0.4138                        & 0.4912                        \\
                                   & \multicolumn{1}{c|}{Hers}                                   & 0.2387                        & 0.3346                        & 0.4586                        & 0.3305                        & 0.4059                        & 0.4842                        \\
                                   & \multicolumn{1}{c|}{GraphRNA}                               & 0.2490                        & 0.3208                        & 0.4372                        & 0.3437                        & 0.4023                        & 0.4755                        \\
                                   & \multicolumn{1}{c|}{ARWMF}                                  & 0.2104                        & 0.3201                        & 0.4512                        & 0.3066                        & 0.3877                        & 0.4704                        \\ \cline{2-8} 
                                   & \multicolumn{1}{c|}{SAT(GCN)-no self}                       & 0.2429                        & 0.3116                        & 0.4614                        & 0.3414                        & 0.3969                        & 0.4889                        \\
                                   & \multicolumn{1}{c|}{SAT(GCN)-no cross}                      & 0.2382                        & 0.3381                        & 0.4611                        & 0.3282                        & 0.4057                        & 0.4835                        \\
                                   & \multicolumn{1}{c|}{SAT(GCN)-no adver}                      & 0.2371                        & 0.3382                        & 0.4707                        & 0.3353                        & 0.4114                        & 0.4966                        \\
                                   & \multicolumn{1}{c|}{SAT(GCN)}                               & 0.2527                        & 0.3560                        & 0.4933                        & 0.3544                        & 0.4332                        & 0.5215                        \\
                                   & \multicolumn{1}{c|}{SAT(GraphSage)}                         & 0.2518                        & 0.3470                        & 0.4845                        & 0.3529                        & 0.4271                        & 0.5133                        \\
\multirow{-14}{*}{Steam}           & \multicolumn{1}{c|}{SAT(GAT)}                               & \textbf{0.2536}               & \textbf{0.3620}               & \textbf{0.4965}               & \textbf{0.3585}               & \textbf{0.4400}               & \textbf{0.5272}               \\ \hline
\end{tabular}
}}
\end{table*}
\subsubsection{Evaluation Measures}
In the \emph{node attribute completion} task, whether the restored attributes can benefit real-world applications should be considerable. Consequently, we propose to measure the quality of restored attributes from both the \emph{node level} and the \emph{attribute level} with two real-world applications.
\begin{itemize}
\setlength{\itemsep}{0pt}
\setlength{\parsep}{0pt}
\setlength{\parskip}{0pt}
 \item Node classification. This task aims to evaluate whether the restored attributes can serve as data augmentation and benefit the classification model. In this task, we use the restored attributes of test nodes to make comparison of node classification among different methods. In other words, this task evaluates the overall quality of restored attributes by classifiers, which is also termed as the evaluation in the \emph{node level}. We implement this task on datasets with class labels.
 \item Profiling. Profile provides a cognitive description for objects such as key terms of papers on Cora and tags of items on Steam. Profiling aims to predict the possible profile for test nodes, and we use Recall@k and NDCG@k as the metrics. In other words, this task evaluates the recall and ranking quality of restored attributes in the \emph{attribute level}. For this task, we compare different methods on datasets with categorical attributes.
\end{itemize}

\begin{figure*}[t]
\centering
\begin{minipage}[t]{0.3\textwidth}
\centering
\includegraphics[width=\textwidth]{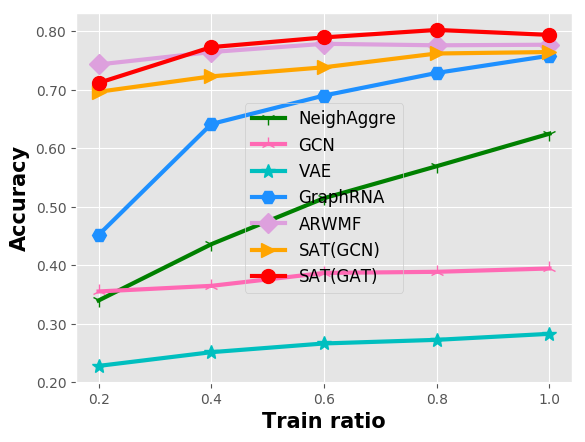}
\caption*{(a) X - Cora}
\end{minipage}
\begin{minipage}[t]{0.3\textwidth}
\centering
\includegraphics[width=\textwidth]{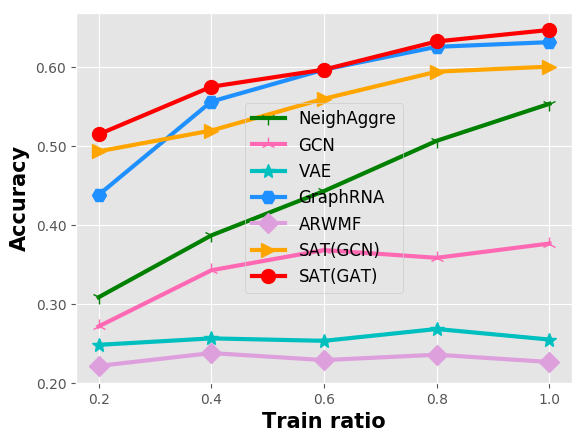}
\caption*{(b) X - Citeseer}
\end{minipage}
\begin{minipage}[t]{0.3\textwidth}
\centering
\includegraphics[width=\textwidth]{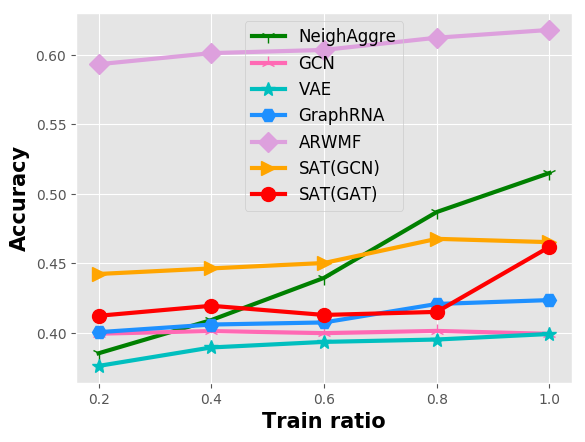}
\caption*{(c) X - Pubmed}
\end{minipage} \\
\begin{minipage}[t]{0.3\textwidth}
\centering
\includegraphics[width=\textwidth]{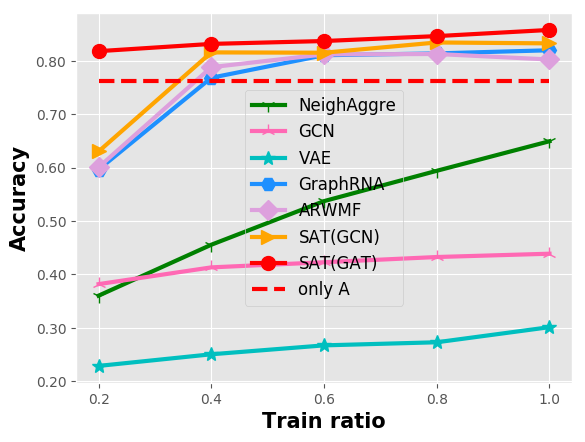}
\caption*{(d) A+X - Cora}
\end{minipage}
\begin{minipage}[t]{0.3\textwidth}
\centering
\includegraphics[width=\textwidth]{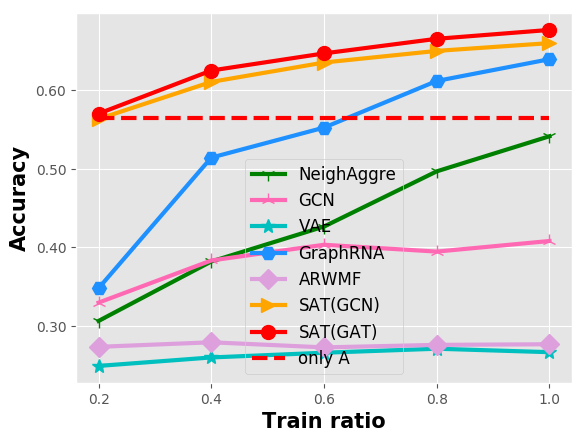}
\caption*{(e) A+X - Citeseer}
\end{minipage}
\begin{minipage}[t]{0.3\textwidth}
\centering
\includegraphics[width=\textwidth]{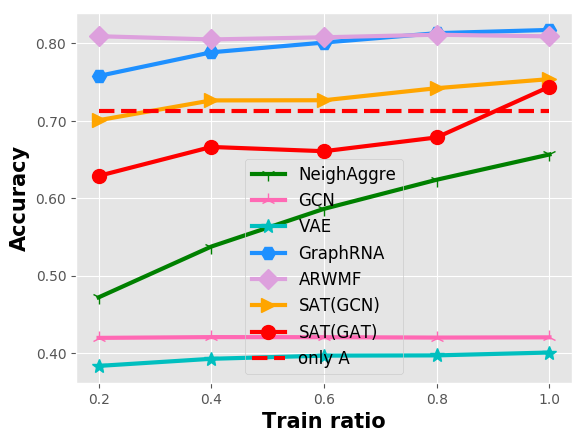}
\caption*{(f) A+X - Pubmed}
\end{minipage} \\
\begin{minipage}[t]{0.3\textwidth}
\centering
\includegraphics[width=\textwidth]{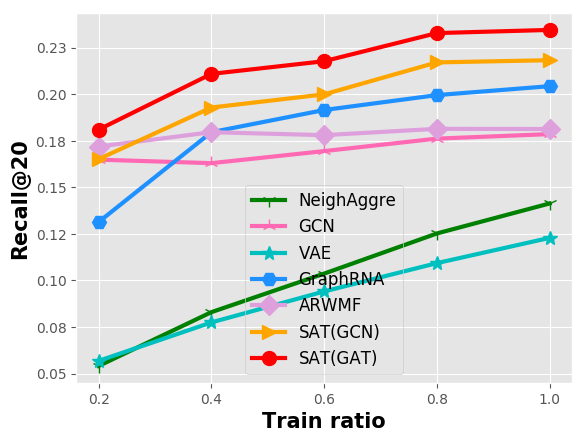}
\caption*{(g) Recall - Cora}
\end{minipage}
\begin{minipage}[t]{0.3\textwidth}
\centering
\includegraphics[width=\textwidth]{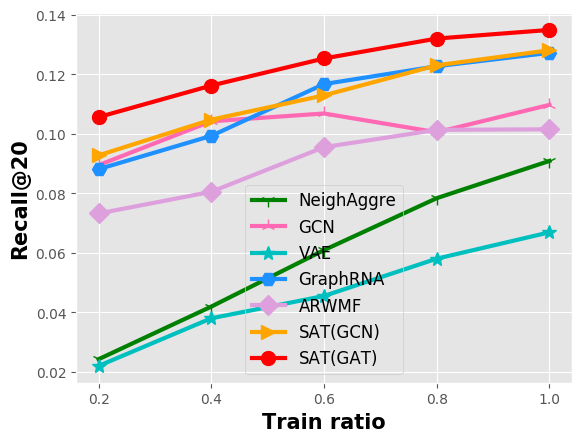}
\caption*{(h) Recall - Citeseer}
\end{minipage}
\begin{minipage}[t]{0.3\textwidth}
\centering
\includegraphics[width=\textwidth]{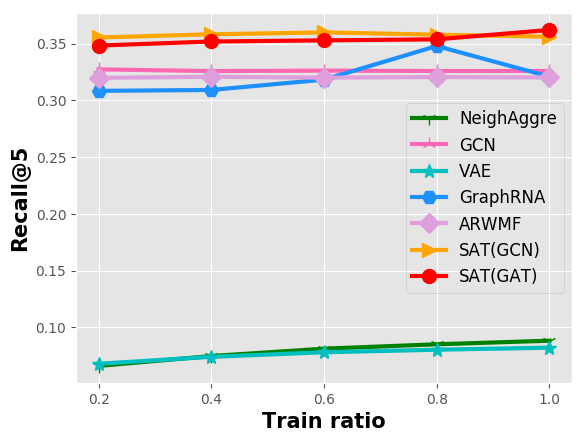}
\caption*{(i) Recall - Steam}
\end{minipage}
\vspace{0pt}
\caption{Node classification and profiling performance with less attribute-observed nodes. (a)-(c) illustrates the results of node classification with ``X'' setting. (d)-(f) shows the results of node classification with ``A+X'' setting. The dashed line is a criterion to criticize whether the restored attributes can enhance the GCN classifier. (g)-(i) shows the results of profiling task. Train ratio means the ratio of samples from the original train data.}
\label{figure:fts_sparsity_res}
\end{figure*}
\subsubsection{Node Classification}
\label{sec:node_classification}
In this task, the restored node attributes are split into 80\% train data and 20\% test data with five-fold validation in 10 times. We consider two classifiers, including MLP and GCN, both with the class information as supervision. Three settings are designed to conduct the comparisons: the \emph{node-attribute-only} approach, the \emph{graph-structure-only} approach, and the \emph{fused} approach. In the \emph{node-attribute-only} approach, we directly use the restored attributes and a two-layer MLP as the classifier to do the classification task. In the \emph{graph-structure-only} approach, only graph structures are applied and this has been studied by many methods such as DeepWalk~\cite{perozzi2014deepwalk}, Node2Vec~\cite{grover2016node2vec} and GCN~\cite{kipf2016semi}. DeepWalk and Node2Vec both aim to learn node embeddings, and then an MLP classifier is used. GCN is an end-to-end method which learns the node embeddings supervised by the classification loss.
In the fused approach, we combine restored node attributes with structures by a GCN classifier.

Table~\ref{table:classification} shows the classification performance, where ``X'' indicates the \emph{node-attribute-only} approach, ``A'' indicates the \emph{structure-only} approach and ``A+X'' is the \emph{fused} one. Considering results of the \emph{node-attribute-only} approach, we have the following observations:
(1) SAT generally presents better performance than baseline methods. For example, compared to NeighAggre, SAT(GCN) reaches nearly 14\% and 5\% gain on Cora and Citeseer, respectively. On Pubmed, it seems that NeighAggre suits this dataset well, but it deteriorates quickly when attribute-observed nodes are less, which is shown in the following experiments. 
(2) The attributed random walk based methods (GraphRNA and ARWMF) show the best performance on Pubmed and Coauthor-CS. It is mainly because the attributed random walk based methods apply random walks on the node-attribute bipartite graphs, which could potentially capture the correlation between attribute dimensions and facilitate the \textit{node attribute completion} task. On Pubmed and Coauthor that have real-valued node attributes, the correlation between attributes could be more obvious because attributes have relative sizes for different attribute dimensions, and thus GraphRNA and ARWMF show better performance. 
Instead, the proposed SAT cannot capture the correlation between attributes and thus perform worse than GraphRNA and ARWMF. 
On other datasets, SAT perform better than GraphRNA and ARWMF mainly because SAT has the advantage of extracting features by graph convolution techniques.
(3) From the results of Hers, we can see that simply taking the graph structures as one kind of side information to augment the recommendation model cannot well exploit the structures and node attributes. And Hers does not perform well on attribute-missing graphs. By contrast, the proposed SAT explicitly models the joint distribution of node structures and attributes and thus shows superior performance compared to Hers.
(4) The performance of SAT(GCN) gets closer to that of true attributes. SAT(GAT) even gets better than the true attributes on Cora mainly because the structural information may take more important role for classification and the restored attributes contain transferred structural information.
(5) Both SAT(GCN)-no self, SAT(GCN)-no cross and SAT(GCN)-no adver perform worse than SAT(GCN) because the incompleteness of SAT cannot guarantee the joint distribution modeling.
(6) SAT with different GNN backbones generally show better performance other baselines, which also states SAT is extendable and robust with different choices of graph convolution modules.

Comparing the results between the \emph{structure-only} approach and the \emph{fused} one, we have the following observations. (1) The restored attributes from SAT(GCN) can augment the GCN classifier with 6.96\%, 9.48\% and 4.12\% gain on Cora, Citeseer and Pubmed, respectively. While NeighAggre fails and harms the GCN performance with the figures as 11.37\%, 2.38\% and 5.61\% on Cora, Citeseer and Pubmed, respectively. (2) The \emph{classical aggregation} method NeighAggre, the \emph{auto-encoding} method VAE, and the \emph{GNN-encoding} methods restore inferior attributes. They hurt the GCN classifier since they are not specifically designed for attribute-missing graphs. 

\begin{table*}[]
\centering
\caption{Area under curve (AUC) and average precision (AP) of Link prediction task. ``A'' indicates the \emph{structure-only} aspect and only structures are used. While ``A+X'' means the \emph{fused structure-attribute} aspect and both structures and attributes are used.}
\label{table:link_prediction_limited}
\renewcommand{\arraystretch}{1.0}
 \setlength{\tabcolsep}{1.0mm}{ 
  \scalebox{1.0}{
\begin{tabular}{cccccccccccccccc}
\hline
\multirow{2}{*}{}                          & \multirow{2}{*}{Method} & \multicolumn{2}{c}{Cora}          & \multicolumn{2}{c}{Citeseer}      & \multicolumn{2}{c}{Pubmed}        & \multicolumn{2}{c}{Steam}         & \multicolumn{2}{c}{\begin{tabular}[c]{@{}c@{}}Coauthor\\ -CS\end{tabular}} & \multicolumn{2}{c}{\begin{tabular}[c]{@{}c@{}}Amazon\\ -Computer\end{tabular}} & \multicolumn{2}{c}{\begin{tabular}[c]{@{}c@{}}Amazon\\ -Photo\end{tabular}} \\ \cline{3-16} 
                                           &                         & AUC             & AP              & AUC             & AP              & AUC             & AP              & AUC             & AP              & AUC                                  & AP                                  & AUC                                    & AP                                    & AUC                                  & AP                                   \\ \hline
\multicolumn{1}{c|}{\multirow{8}{*}{A}}    & SPM                     & 0.8627          & 0.9001          & 0.8293          & 0.8646          & 0.8339          & 0.8544          & 0.8848          & 0.9355          & 0.9169                               & 0.9385                              & 0.9345                                 & 0.9309                                & 0.9372                               & \textbf{0.9416}                      \\
\multicolumn{1}{c|}{}                      & DeepWalk                & 0.8735          & 0.9018          & 0.8082          & 0.8467          & 0.8361          & 0.8560          & 0.9591          & 0.9217          & 0.9174                               & 0.9119                              & 0.8735                                 & 0.8657                                & 0.9148                               & 0.9079                               \\
\multicolumn{1}{c|}{}                      & Node2Vec                & 0.8496          & 0.8695          & 0.7644          & 0.8246          & 0.8346          & 0.8884          & 0.9591          & 0.9233          & 0.9359                               & 0.9348                              & 0.8551                                 & 0.8355                                & 0.8770                               & 0.8608                               \\
\multicolumn{1}{c|}{}                      & GAE(GCN)                & 0.8281          & 0.8309          & 0.8254          & 0.8344          & 0.7072          & 0.6548          & 0.9804          & 0.9772          & 0.7674                               & 0.7307                              & 0.8436                                 & 0.7755                                & 0.9135                               & 0.8960                               \\
\multicolumn{1}{c|}{}                      & GAE(GAT)                & 0.8340          & 0.8642          & 0.8563          & 0.8777          & 0.7028          & 0.6559          & 0.9730          & 0.9637          & 0.7702                               & 0.7359                              & 0.8408                                 & 0.8419                                & 0.8802                               & 0.8493                               \\
\multicolumn{1}{c|}{}                      & VGAE(GCN)               & 0.8433          & 0.8616          & 0.8396          & 0.8456          & 0.8011          & 0.8171          & 0.9738          & 0.9693          & 0.8591                               & 0.8440                              & 0.9345                                 & 0.9281                                & 0.9359                               & 0.9329                               \\
\multicolumn{1}{c|}{}                      & VGAE(GAT)               & 0.7267          & 0.7329          & 0.7046          & 0.7156          & 0.7011          & 0.6264          & 0.8227          & 0.8094          & 0.7370                               & 0.6935                              & 0.7492                                 & 0.6564                                & 0.8313                               & 0.7500                               \\
\multicolumn{1}{c|}{}                      & Hers                    & 0.6908          & 0.7169          & 0.6779          & 0.7197          & 0.5001          & 0.5182          & 0.9869          & 0.9818          & 0.5433                               & 0.5338                              & 0.5314                                 & 0.5122                                & 0.5809                               & 0.5589                               \\ \hline
\multicolumn{1}{c|}{\multirow{11}{*}{A+X}} & GAE(GCN)                & 0.8375          & 0.8374          & 0.8227          & 0.8245          & 0.7947          & 0.7855          & 0.9735          & 0.9643          & 0.9409                               & \textbf{0.9386}                     & 0.9361                                 & 0.9336                                & 0.9336                               & 0.9296                               \\
\multicolumn{1}{c|}{}                      & GAE(GAT)                & 0.8575          & 0.8493          & 0.8387          & 0.8661          & 0.7839          & 0.7821          & 0.9717          & 0.9706          & \textbf{0.9452}                      & 0.9374                              & 0.7945                                 & 0.7836                                & 0.9441                               & 0.9282                               \\
\multicolumn{1}{c|}{}                      & VGAE(GCN)               & 0.7910          & 0.8018          & 0.7962          & 0.8011          & 0.8229          & 0.8156          & 0.9581          & 0.9454          & 0.8960                               & 0.8936                              & 0.9315                                 & 0.924                                 & 0.9205                               & 0.9202                               \\
\multicolumn{1}{c|}{}                      & VGAE(GAT)               & 0.7062          & 0.7172          & 0.7069          & 0.7061          & 0.7861          & 0.7578          & 0.9728          & 0.9195          & 0.7678                               & 0.7382                              & 0.7609                                 & 0.6909                                & 0.7125                               & 0.6699                               \\
\multicolumn{1}{c|}{}                      & GraphRNA                & 0.8579          & 0.8695          & 0.8298          & 0.8580          & 0.8406          & 0.8445          & 0.9157          & 0.8807          & 0.8688                               & 0.8579                              & 0.7115                                 & 0.6859                                & 0.8359                               & 0.8010                               \\
\multicolumn{1}{c|}{}                      & ARWMF                   & 0.7601          & 0.7770          & 0.7260          & 0.7646          & 0.7022          & 0.7320          & 0.7832          & 0.6580          & 0.8582                               & 0.8503                              & 0.8325                                 & 0.8251                                & 0.9102                               & 0.8949                               \\ \cline{2-16} 
\multicolumn{1}{c|}{}                      & SAT(GCN)-no self        & 0.8549          & 0.8432          & 0.8542          & 0.8550          & 0.7661          & 0.7497          & 0.9702          & 0.9599          & 0.8166                               & 0.8029                              & 0.8593                                 & 0.8253                                & 0.8888                               & 0.8470                               \\
\multicolumn{1}{c|}{}                      & SAT(GCN)-no cross       & 0.8048          & 0.7894          & 0.8042          & 0.8097          & 0.7817          & 0.7998          & 0.9617          & 0.9560          & 0.7990                               & 0.7592                              & 0.8877                                 & 0.8918                                & 0.8950                               & 0.8556                               \\
\multicolumn{1}{c|}{}                      & SAT(GCN)-no adver       & 0.8438          & 0.8405          & 0.8438          & 0.8459          & 0.7851          & 0.8018          & 0.9703          & 0.9605          & 0.9005                               & 0.8920                              & 0.9363                                 & 0.9303                                & 0.9466                               & 0.9390                               \\
\multicolumn{1}{c|}{}                      & SAT(GCN)                & 0.8554          & 0.8500          & 0.8569          & 0.8566          & 0.8258          & 0.8039          & 0.9713          & 0.9636          & 0.9077                               & 0.8977                              & \textbf{0.9433}                        & \textbf{0.9368}                       & \textbf{0.9467}                      & 0.9393                               \\
\multicolumn{1}{c|}{}                      & SAT(GAT)                & \textbf{0.8929} & \textbf{0.9018} & \textbf{0.8916} & \textbf{0.9139} & \textbf{0.8510} & \textbf{0.8554} & \textbf{0.9896} & \textbf{0.9881} & 0.9138                               & 0.9035                              & 0.9100                                 & 0.8940                                & 0.9283                               & 0.9112                               \\ \hline
\end{tabular}
}}
\end{table*}

\subsubsection{Profiling}
For the profiling task, restored attributes are probabilities that the node may have in each attribute dimension. High-quality restored attributes should have high probabilities in specific dimensions as the true attributes. Taking the recall and ranking ability into consideration, we use Recall@k and NDCG@k to evaluate \emph{node attribute completion} in the \emph{attribute level}. The results are shown in Table~\ref{table:label_propagation}.

From Table~\ref{table:label_propagation}, we can summarize as follows. (1) NeighAggre performs the worst among all methods on the three datasets. It is not a learning algorithm and cannot perform robustly in this fine-grained \emph{attribute-level} evaluation. 
(3) The attributed random walk based methods have the potential to capture the correlation between attribute dimensions, GraphRNA and ARWMF show the most competitive performance on this \emph{attribute-level} evaluation. However, the applied random walks may introduce noise to statistically represent the original graph data. Thereby, GraphRNA and ARWMF cannot outperform SAT on this fine-grained \emph{attribute-level} evaluation.
(4) SAT achieves superior performance over other methods because it restores attributes based on the transformation knowledge.

\subsubsection{Less Attribute-Observed Nodes}\label{sec:less_nodes}
The attribute-observed nodes are necessary and supervise the learning on attribute-missing graphs. In some scenarios, this supervision could be less, so it is curious to see whether SAT can still restore reliable and high-quality node attributes when less attribute-observed nodes are available. We experiment to see the node classification and profiling performance in this scenario.

Figure~\ref{figure:fts_sparsity_res} (a)(b)(c) show the node classification results when only node attributes are used. From these figures, we can see that SAT(GCN) generally performs much better than other methods. NeighAggre deteriorates quickly and the gap is more obvious when attribute-observed nodes are less.
Figure~\ref{figure:fts_sparsity_res} (d)(e)(f) show the node classification performance when ``A+X'' is used. In these figures, the dotted line indicates only ``A'' is used by a GCN classifier, which serves as a criterion to evaluate whether the restored attributes can enhance the GCN classifier. It is clear that SAT(GCN) reaches superior performance than baselines on Cora and Citeseer. On Pubmed with real-valued node attributes, GraphRNA and ARWMF show better performance because they have the potential to capture the correlation between attribute dimensions. Figure~\ref{figure:fts_sparsity_res} (g)(h)(i) indicate the fine-grained profiling performance. These figures also demonstrate the advantage of SAT in restoring fine-grained node attributes.

\subsection{Link Prediction}
\subsubsection{Overall Performance}
Link prediction is a key problem on graphs which aims to predict the missing links among nodes. It has various applications such as friend recommendation~\cite{adamic2003friends}. We conduct the link prediction task on attribute-missing graphs from two aspects: the \emph{structure-only} aspect and the \emph{fused structure-attribute} aspect. The results are shown in Table~\ref{table:link_prediction_limited}.

In Table~\ref{table:link_prediction_limited}, ``A'' indicates the \emph{structure-only} aspect where only structures are used. While ``A+X'' means the \emph{fused structure-attribute} aspect where both structures and attributes are used. From this table, we can summarize that: (1) SPM with the idea of perturbations on graphs show competitive performance because of its denoising characteristic. 
However, it is limited in learning deep representations for nodes and cannot beat SAT. It is also worthwhile to point out that SAT does not have the denoising characteristic and introducing the perturbation idea from SPM to SAT is also a promising direction for link prediction task. 
(3) In the ``A+X'' setting, SAT generally outperforms recent GNN based methods (e.g. GAE and VGAE) and attributed random walk based methods (e.g. GraphRNA and ARWMF). Compared to GAE(GAT), SAT(GAT) obtains a 3.73\% gain, 6.2\% gain and 6.71\% gain on Cora, Citeseer and Pubmed respectively. This also indicates that SAT has the advantage of using attribute-observed nodes while recent GNN methods cannot handle the link prediction task in this scenario properly. Compared to GraphRNA and ARWMF, SAT has the advantage of encoding structures with graph convolution scheme and thus show better performance on link prediction task.
(4) Breaking the unified loss in Eq.~\ref{eq:objective} causes deterioration to the performance since it cannot guarantee the joint distribution modeling depicted in Section~\ref{section:method_formulation}.
\begin{figure*}[t]
\centering
\begin{minipage}[t]{0.32\textwidth}
\centering
\includegraphics[width=\textwidth]{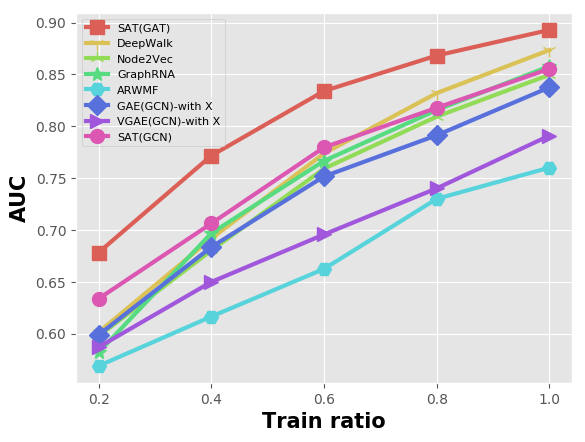}
\caption*{(a) Cora}
\end{minipage}
\begin{minipage}[t]{0.32\textwidth}
\centering
\includegraphics[width=\textwidth]{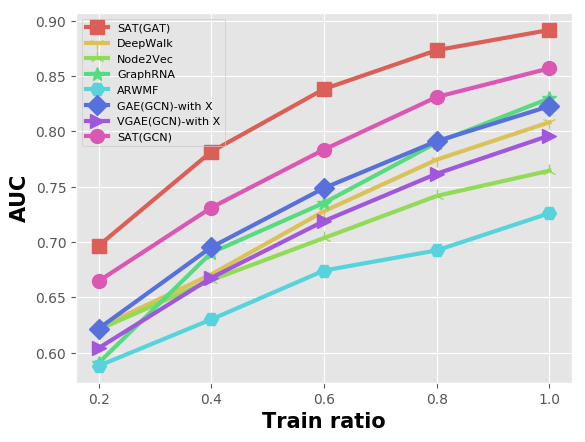}
\caption*{(b) Citeseer}
\end{minipage}
\begin{minipage}[t]{0.32\textwidth}
\centering
\includegraphics[width=\textwidth]{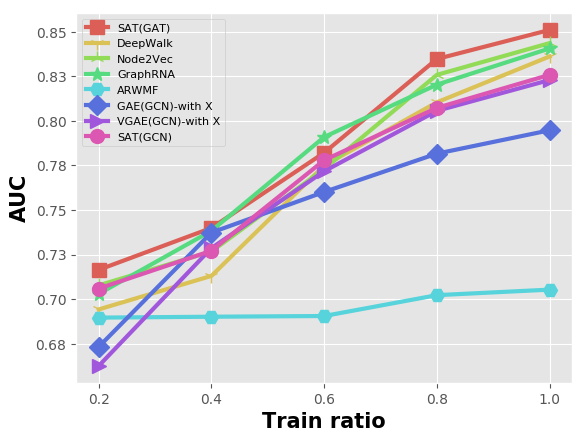}
\caption*{(c) Pubmed}
\end{minipage} 
\vspace{0pt}
\caption{Link prediction with less observed links on three datasets. We use AUC here to evaluate the performance. Train ratio means the ratio of random samples from original train data. Term ``with X'' means attributes are taken into consideration and ``no X'' means only structures are considered.}
\label{figure:link_sparsity4link_prediction}
\end{figure*}

\begin{figure*}[]
\centering
\begin{minipage}[t]{0.23\textwidth}
\centering
\includegraphics[width=\textwidth]{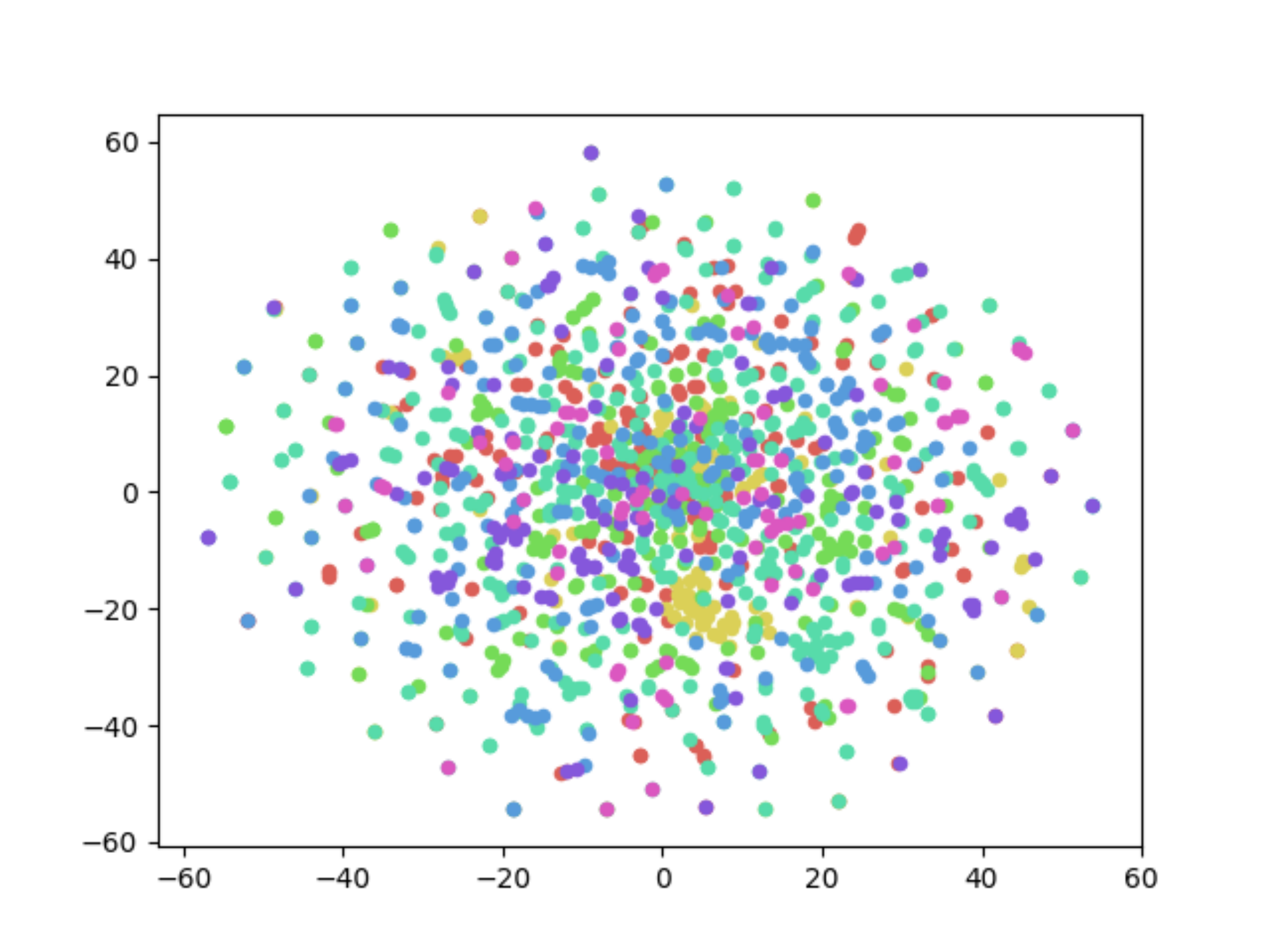}
\caption*{(a) VAE}
\end{minipage}
\begin{minipage}[t]{0.23\textwidth}
\centering
\includegraphics[width=\textwidth]{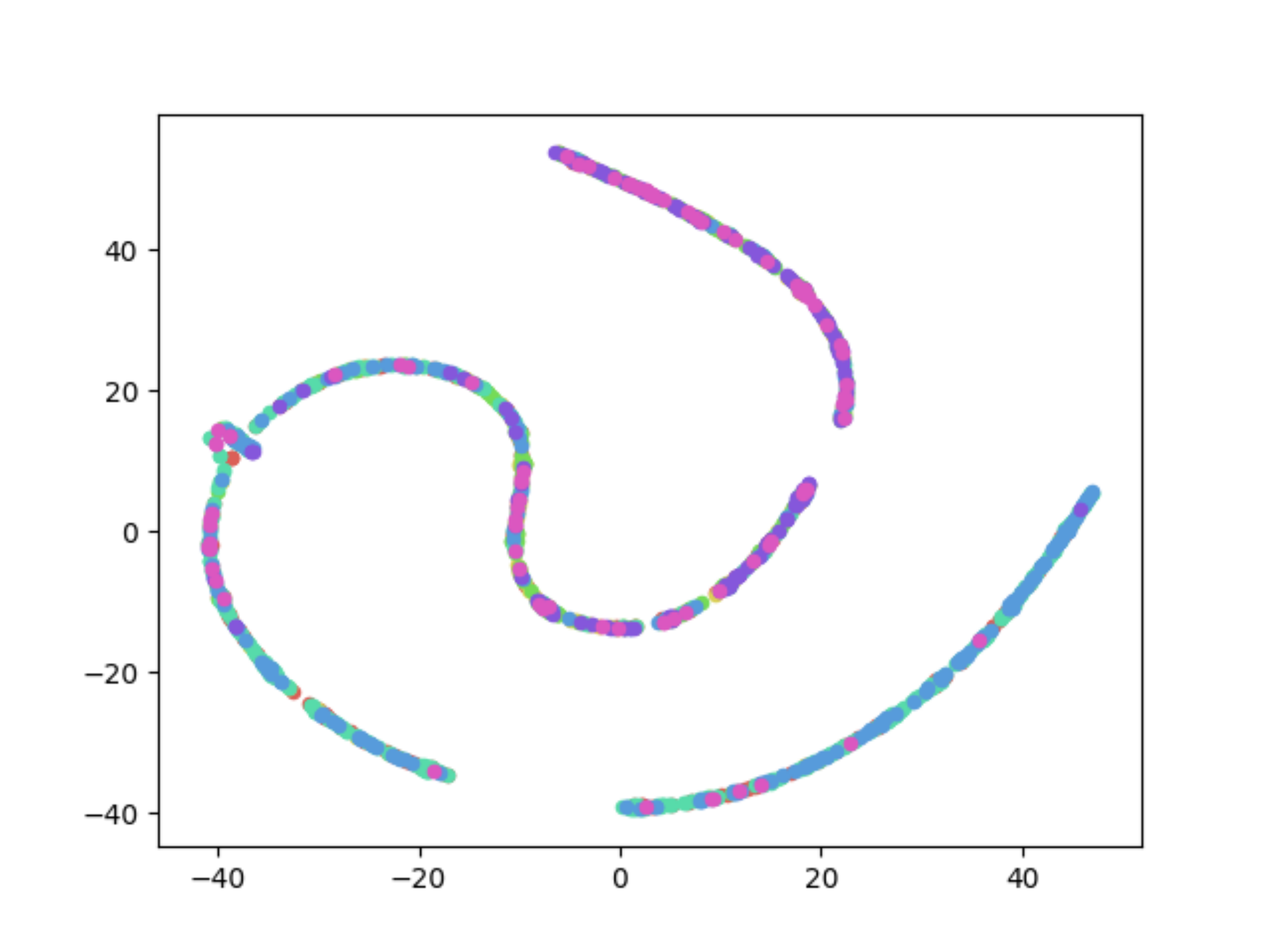}
\caption*{(b) GCN}
\end{minipage}
\begin{minipage}[t]{0.23\textwidth}
\centering
\includegraphics[width=\textwidth]{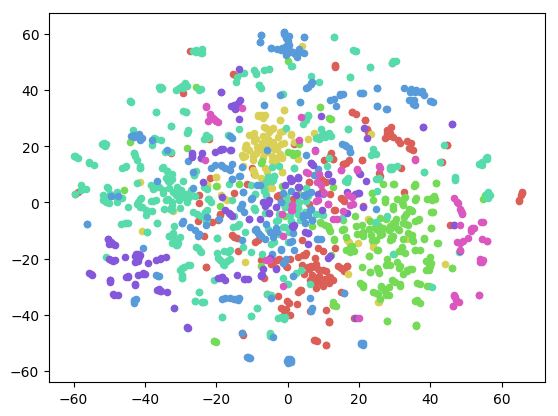}
\caption*{(c) GraphRNA}
\end{minipage}
\begin{minipage}[t]{0.23\textwidth}
\centering
\includegraphics[width=\textwidth]{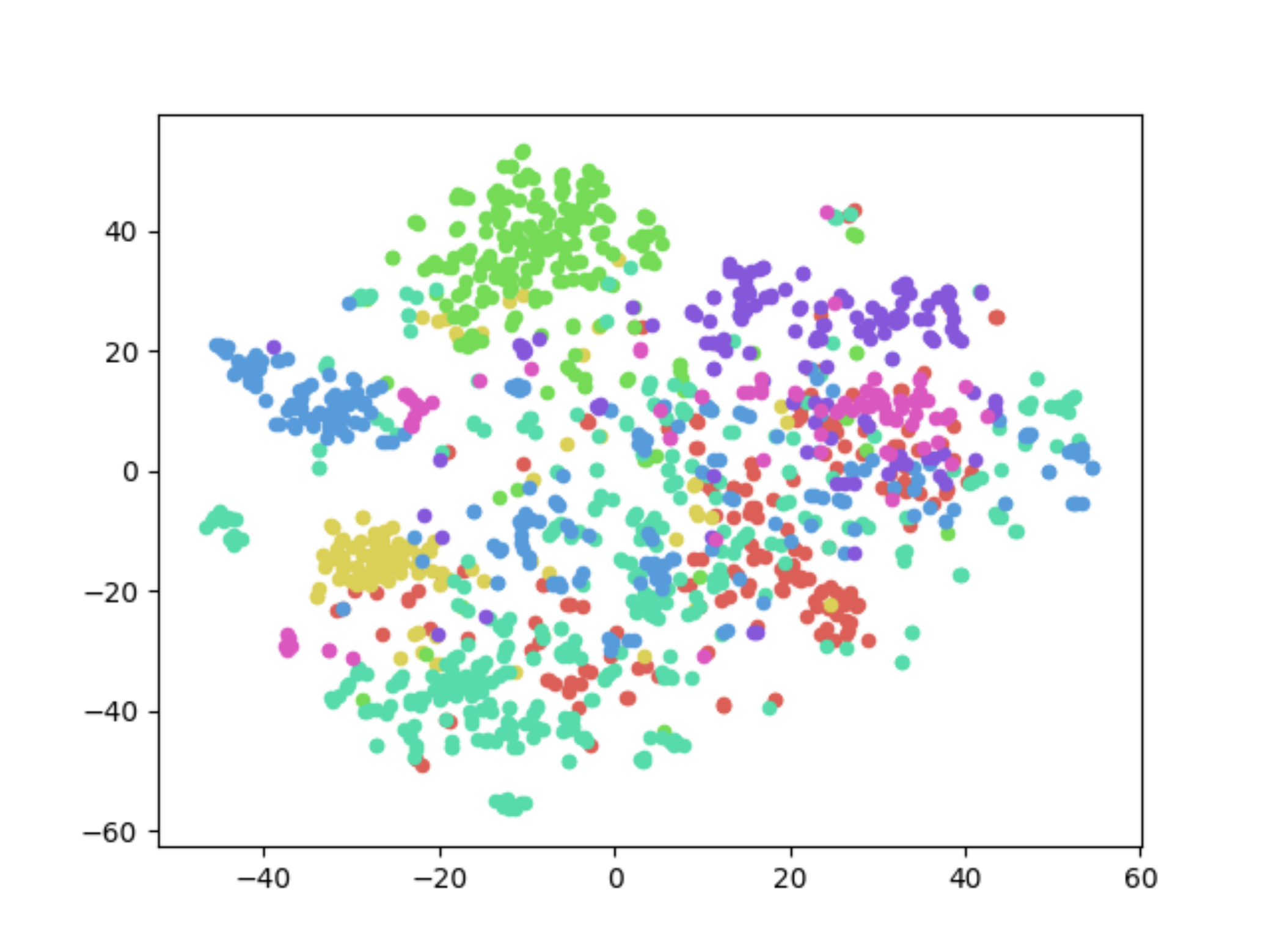}
\caption*{(d) SAT(GCN)}
\end{minipage}
\caption{Learned node representation analysis. The t-SNE visualization of test node embeddings on Cora. Each color represents one class. Note that all methods learn node embeddings without class supervision.}
\label{figure:TSNE_result}
\end{figure*}
\subsubsection{Less Observed Links}
In link prediction task, the robustness of methods with less observed links is usually explored. We also conduct the link prediction experiment of link sparsity here. The results are shown in Figure~\ref{figure:link_sparsity4link_prediction}.

From Figure~\ref{figure:link_sparsity4link_prediction}, we can summarize that: (1) SAT(GAT) outperforms other competitive baselines even with less observed links. Considering the results of SAT(GCN) and SAT(GAT), we see that GAT is a strong graph convolution module and benefits SAT a lot. 
(2) Applying zero-filling trick for attribute-missing nodes sometimes augments the models (e.g. GAE) but it is not consistent on all datasets. (3) On Pubmed, DeepWalk and Node2Vec are the most two competitive baselines, they even beat SAT(GCN) in some sparse cases. Introducing the advantageous ranking loss from DeepWalk and Node2Vec to GNN might augment SAT(GCN) for link prediction task, but we do not explore it here since we mainly focus on the general method for learning on attribute-missing graphs. 

\subsection{Learned Node Representation Analysis}
\begin{figure*}[ht]
\centering
\begin{minipage}[t]{0.3\textwidth}
\centering
\includegraphics[width=\textwidth]{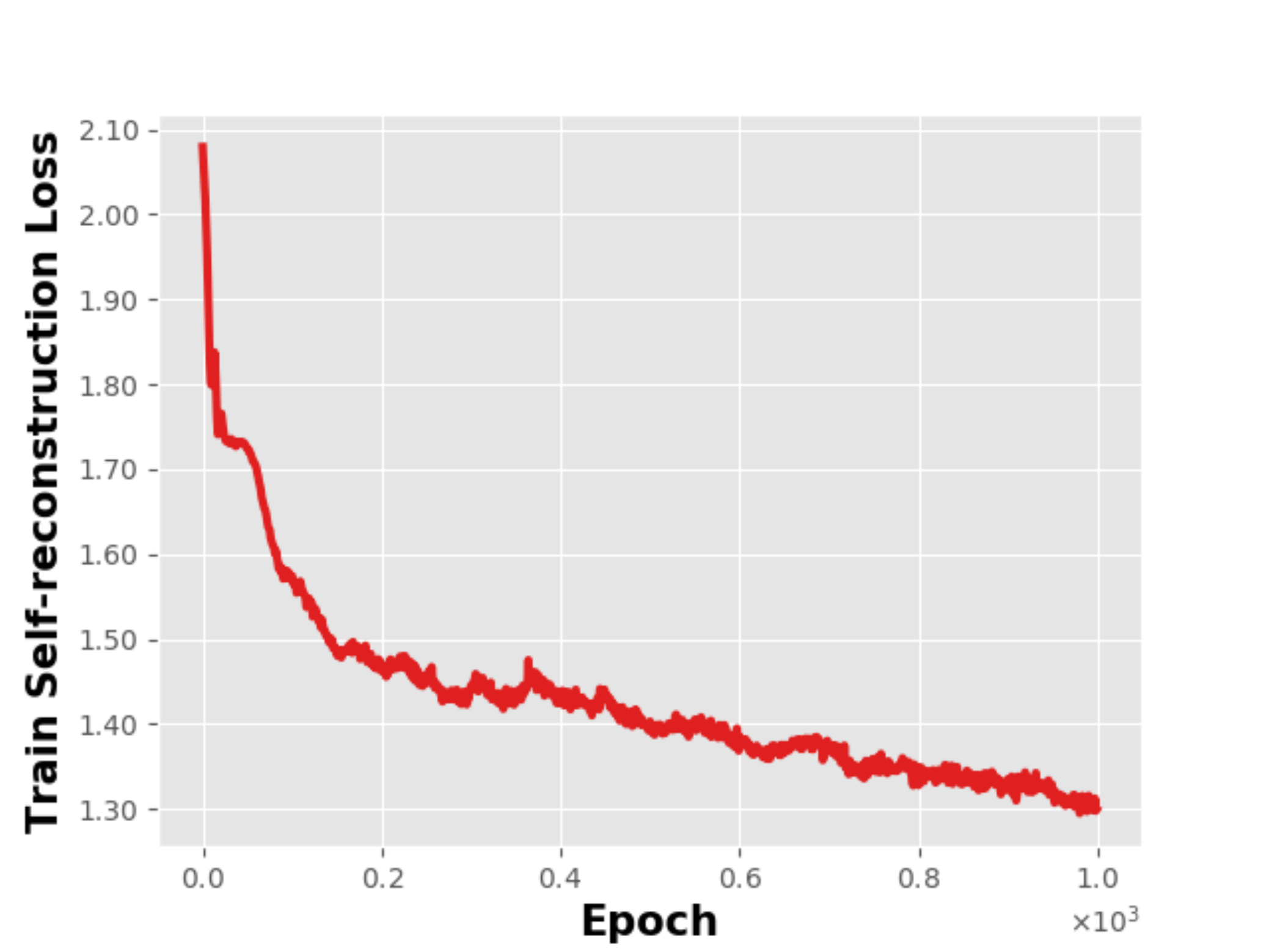}
\caption*{(a) Train self-reconstruction loss}
\end{minipage}
\begin{minipage}[t]{0.3\textwidth}
\centering
\includegraphics[width=\textwidth]{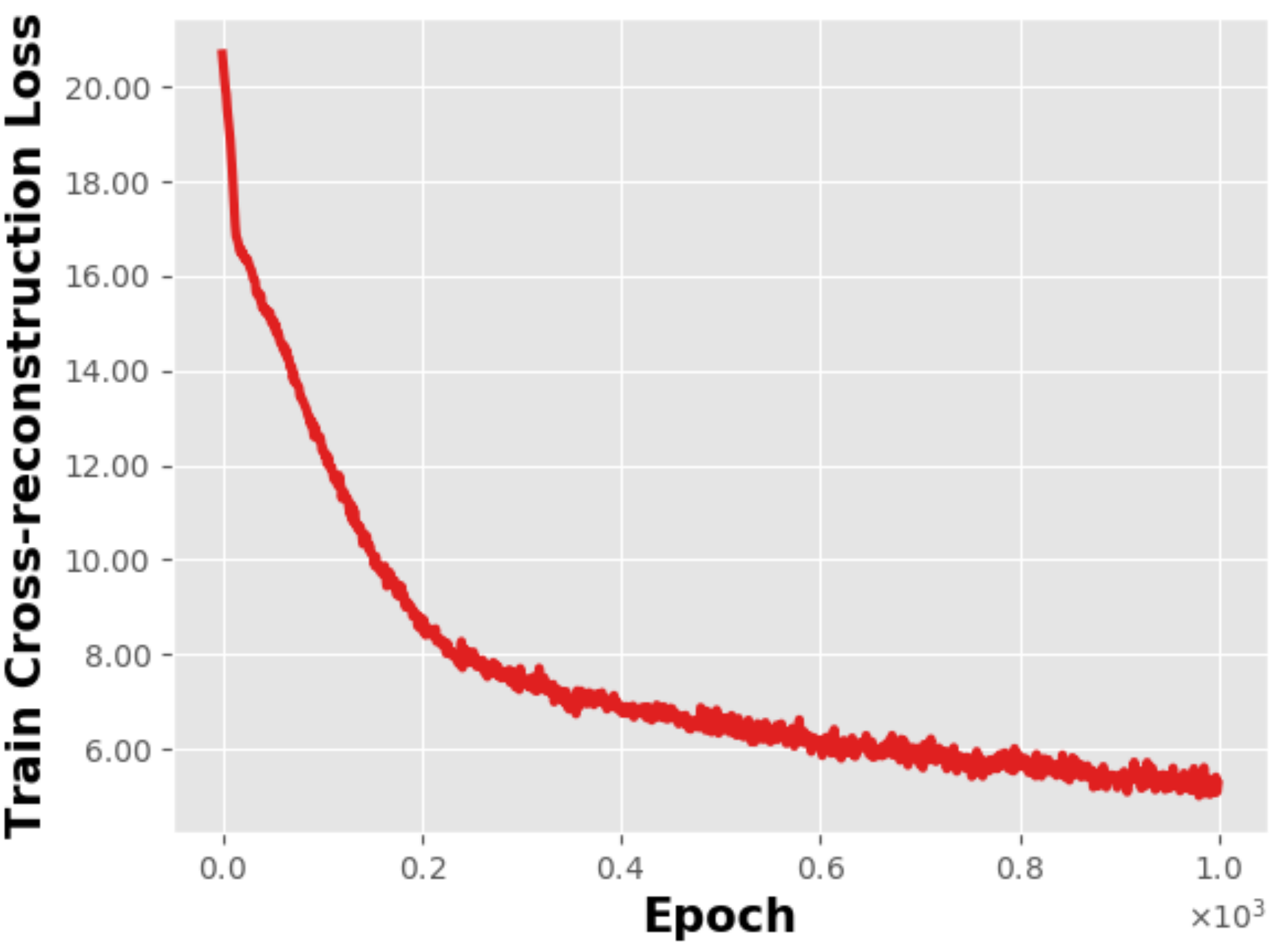}
\caption*{(b) Train cross-reconstruction loss}
\end{minipage}
\begin{minipage}[t]{0.3\textwidth}
\centering
\includegraphics[width=\textwidth]{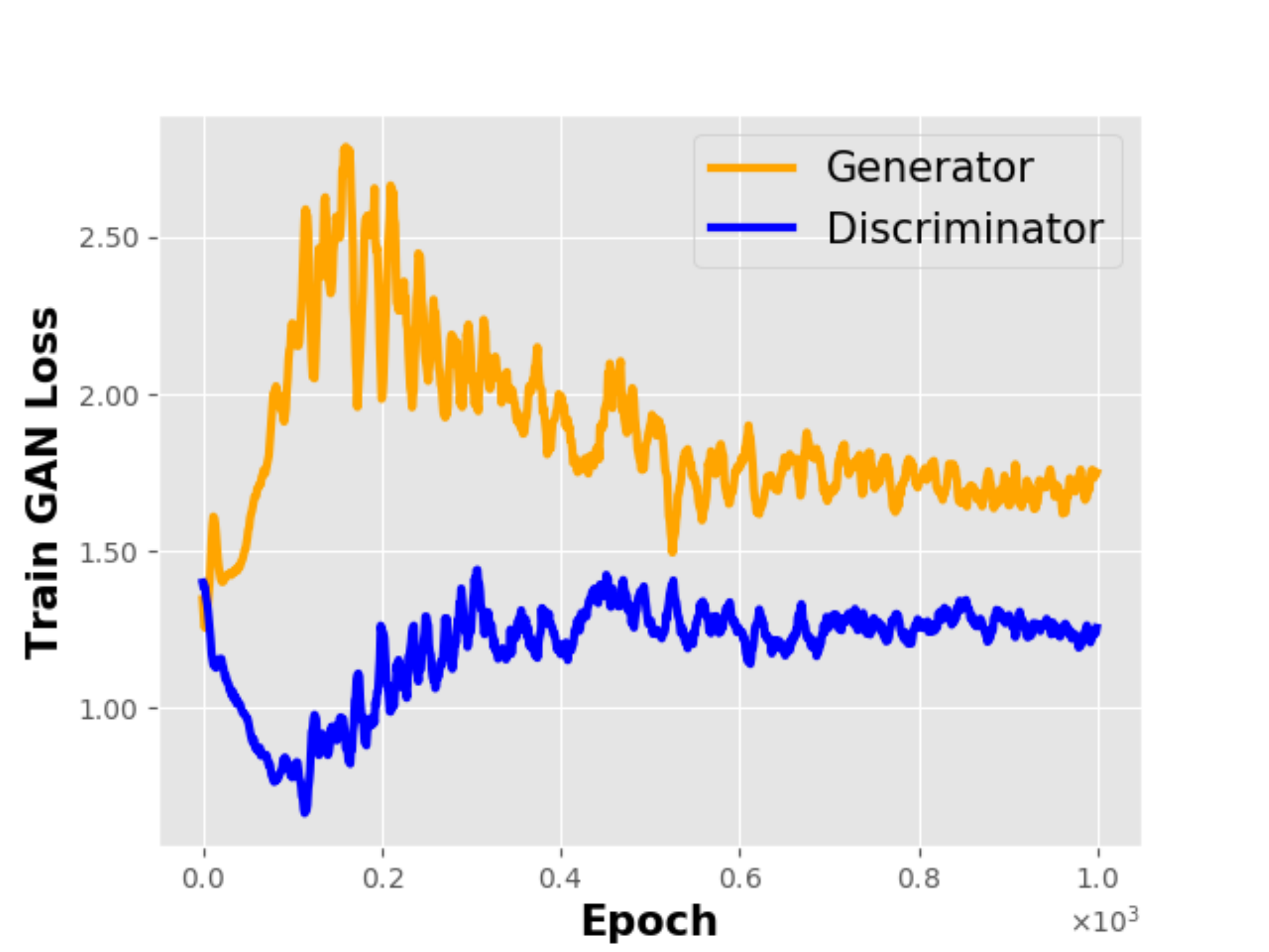}
\caption*{(c) Train GAN loss}
\end{minipage} \\
\begin{minipage}[t]{0.3\textwidth}
\centering
\includegraphics[width=\textwidth]{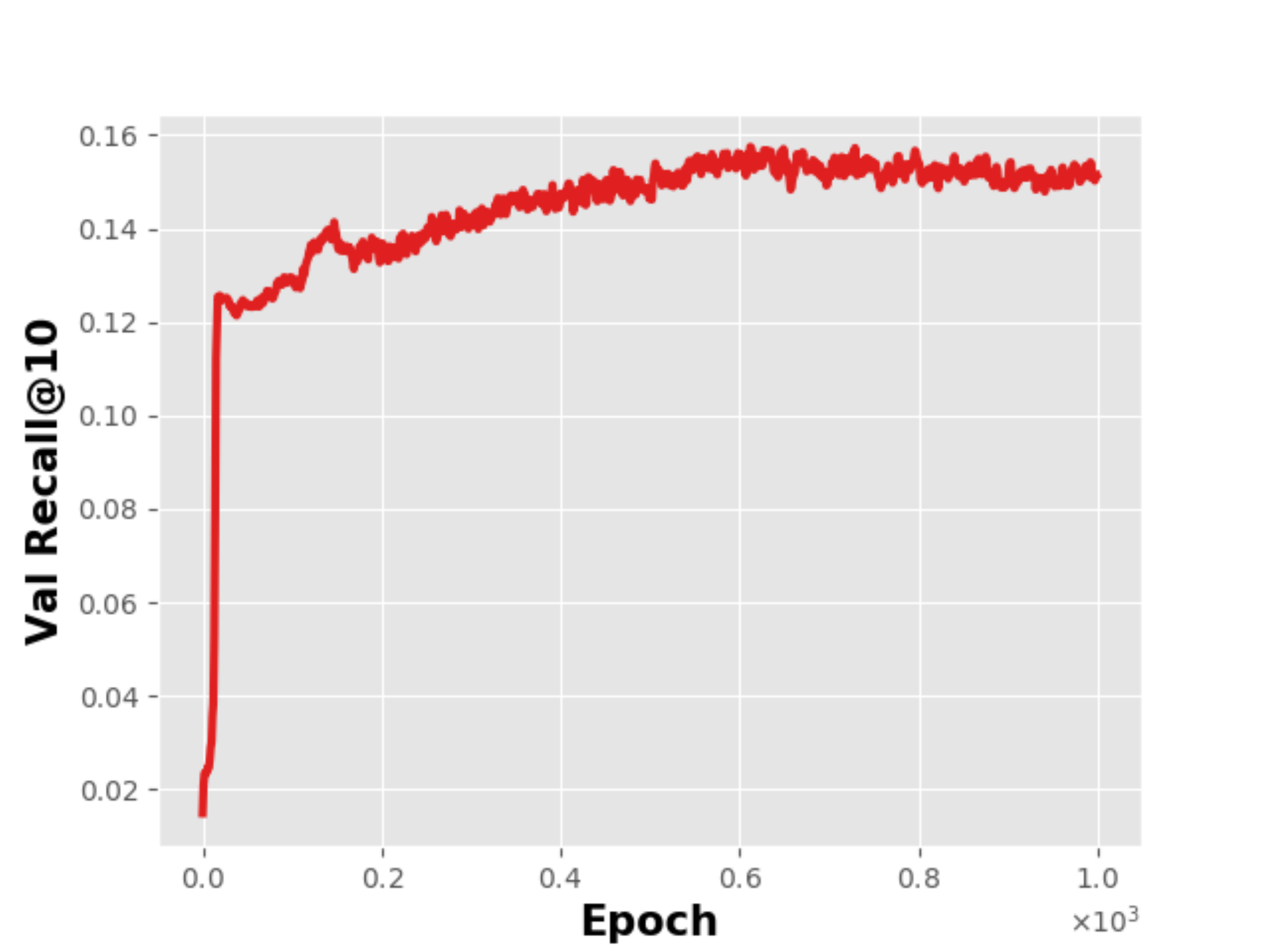}
\caption*{(d) Validation metric}
\end{minipage} 
\begin{minipage}[t]{0.3\textwidth}
\centering
\includegraphics[width=\textwidth]{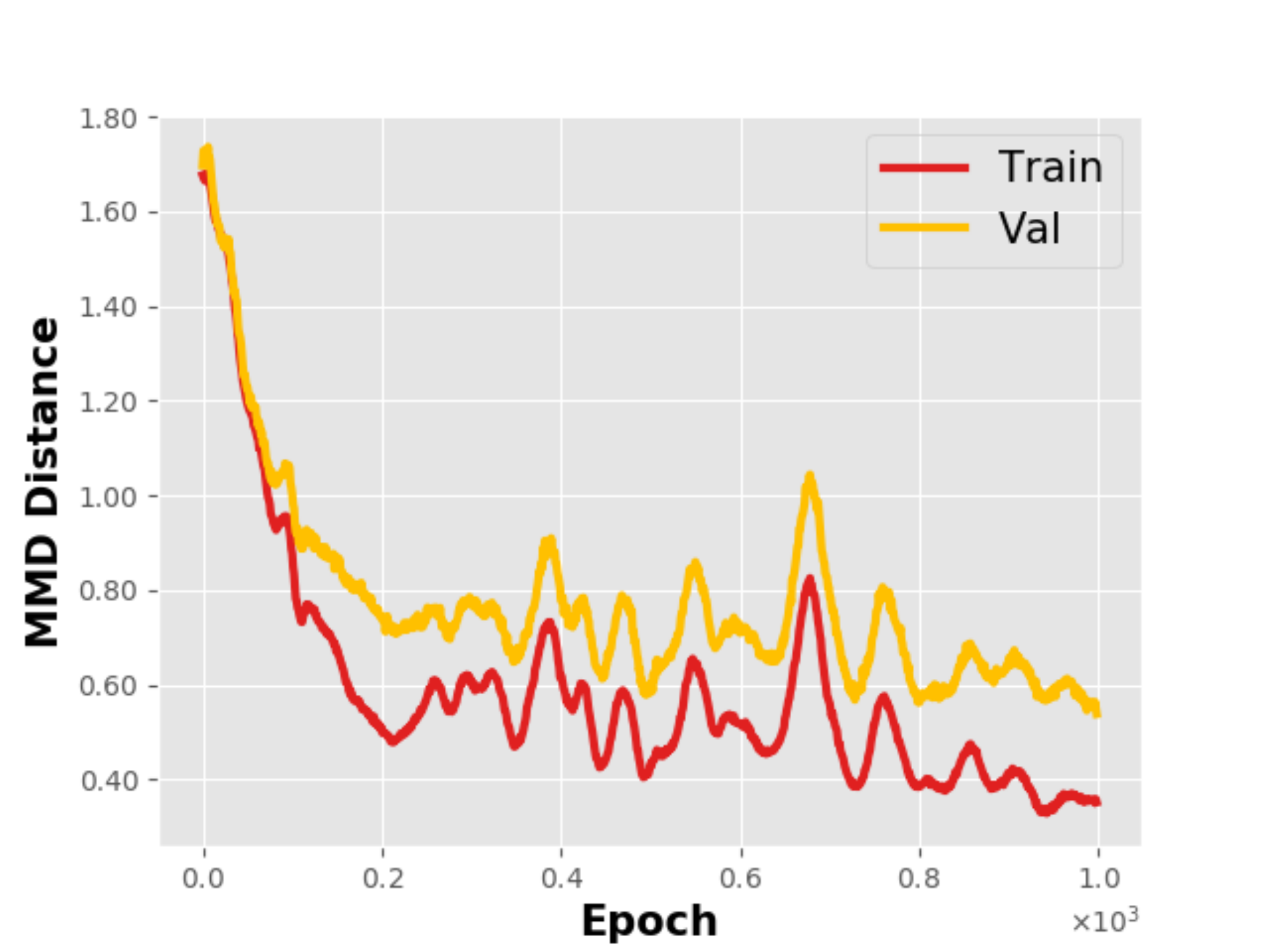}
\caption*{(e) MMD distance}
\end{minipage}
\begin{minipage}[t]{0.3\textwidth}
\centering
\includegraphics[width=\textwidth]{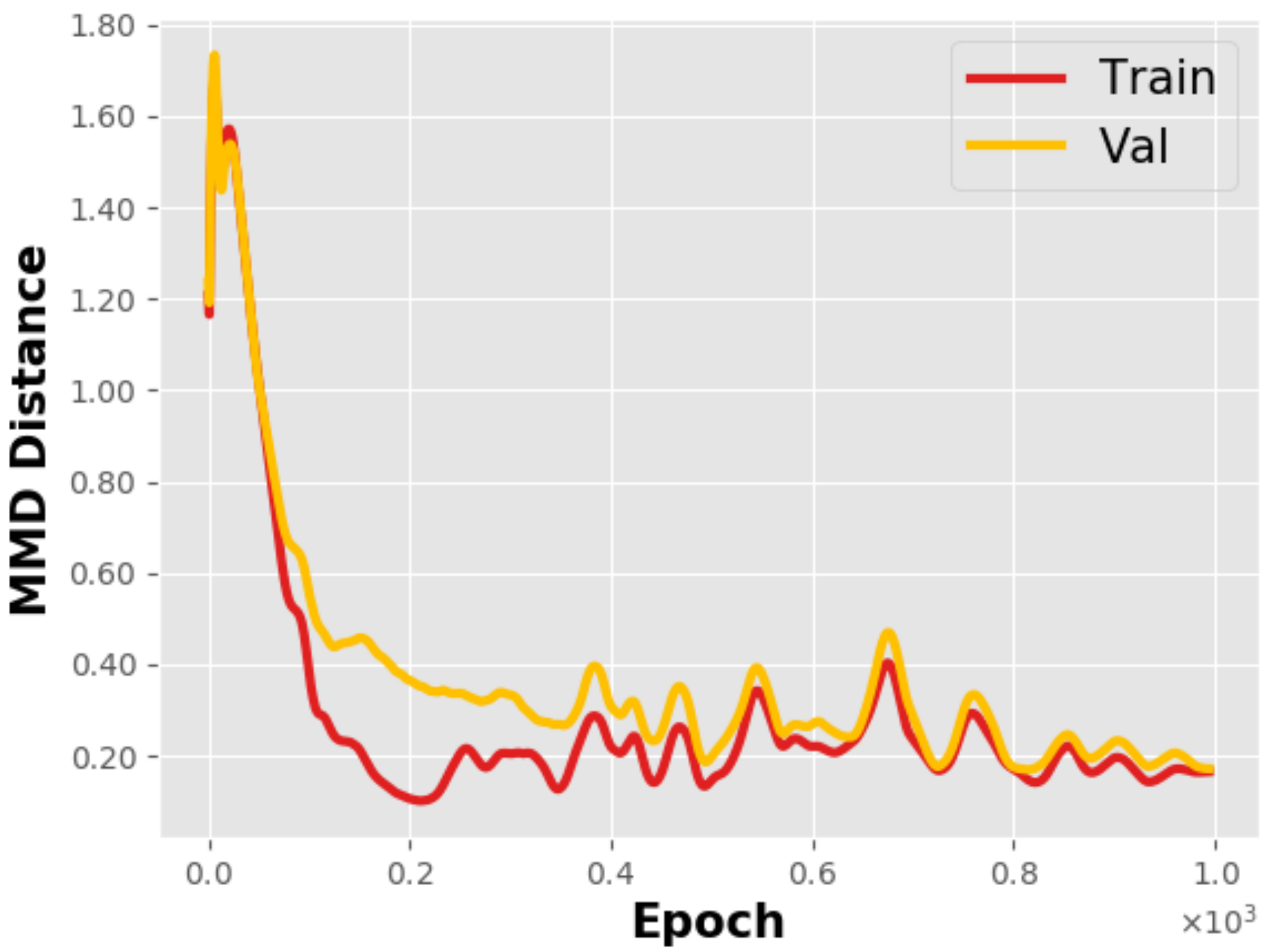}
\caption*{(f) MMD distance}
\end{minipage}
\caption{Visualization of the training process for SAT(GCN) on Cora. (a) The self-reconstruction loss. (b) The cross-reconstruction loss. (c) The GAN loss in adversarial distribution matching. (d) Validation Recall@10 along the training steps. (d) The train and validation MMD distance between the aggregated distribution $q(z)$ and Gaussian prior $p(z)$. (e) The train and validation MMD distance between distributions of $z_{x}$ and $z_{a}$.}
\label{figure:Train_process}
\end{figure*}
In \emph{node attribute completion}, the involved methods, VAE, GCN, GraphRNA and SAT(GCN) infer node attributes by learning the latent embeddings of nodes. Good representation ability means that a method can learn representative embeddings where nearby nodes correspond to similar objects. Therefore, we experiment to visualize the learned node embeddings by t-SNE. Specifically, the latent embeddings for all test nodes are sampled. Then we use t-SNE to make dimension reduction and visualize them in 2-D space on Cora dataset. Nodes in the same class are expected to be clustered together. Note that for all methods, they do not use any label information in the training process. Therefore, the embeddings for t-SNE visualization are learned in an unsupervised manner. Figure~\ref{figure:TSNE_result} shows the results.

From Figure~\ref{figure:TSNE_result} (a) of VAE, we can see that the nodes of different classes are mixed, which means VAE cannot distinguish the nodes belonging to different categories. For GCN in Figure~\ref{figure:TSNE_result} (b), the nodes are encoded into a narrow and stream-like space, where different nodes are mixed and overlapped. Compared to VAE, GCN has no prior assumption, which makes it lose distributed constraint and lead to the narrow and stream space. Compared to GraphRNA in Figure
As for our SAT(GCN) in ~\ref{figure:TSNE_result} (c), SAT(GCN) shows more distinguished node representations and different nodes are clustered well. Although Gaussian prior is imposed on the latent space of both VAE and SAT, our SAT can capture the joint relationship between attributes and structures, and learn better latent codes for nodes whose attributes are missing.
\begin{figure*}[]
\centering
\begin{minipage}[t]{0.3\textwidth}
\centering
\includegraphics[width=\textwidth]{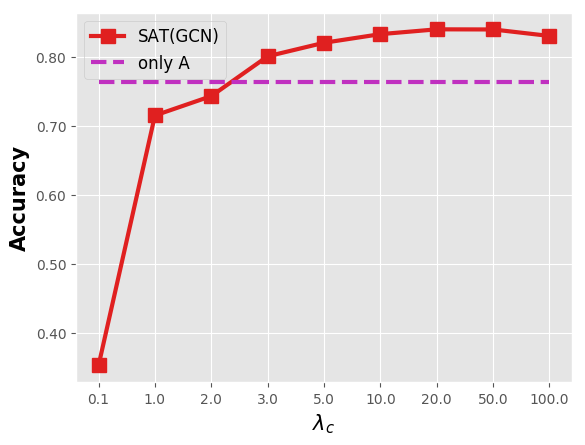}
\caption*{(a) A+X - Cora}
\end{minipage}
\begin{minipage}[t]{0.3\textwidth}
\centering
\includegraphics[width=\textwidth]{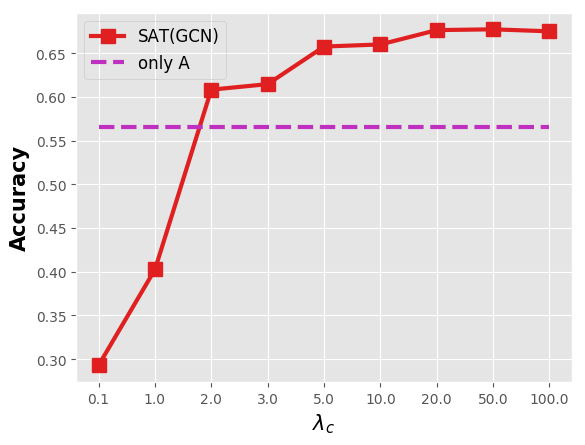}
\caption*{(b) A+X - Citeseer}
\end{minipage}
\begin{minipage}[t]{0.3\textwidth}
\centering
\includegraphics[width=\textwidth]{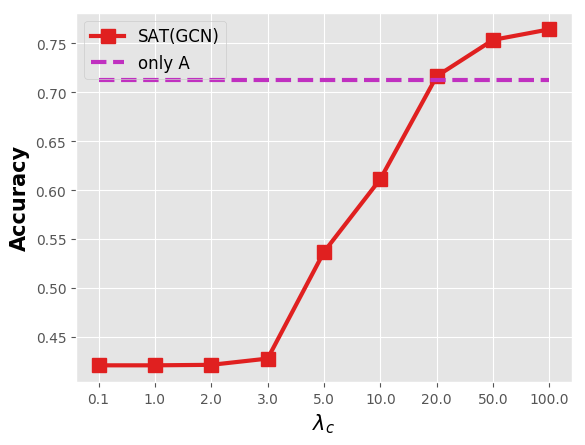}
\caption*{(c) A+X - Pubmed}
\end{minipage} \\
\begin{minipage}[t]{0.3\textwidth}
\centering
\includegraphics[width=\textwidth]{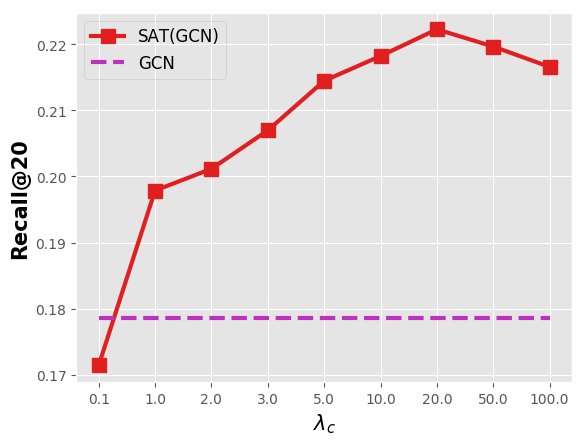}
\caption*{(d) Recall - Cora}
\end{minipage}
\begin{minipage}[t]{0.3\textwidth}
\centering
\includegraphics[width=\textwidth]{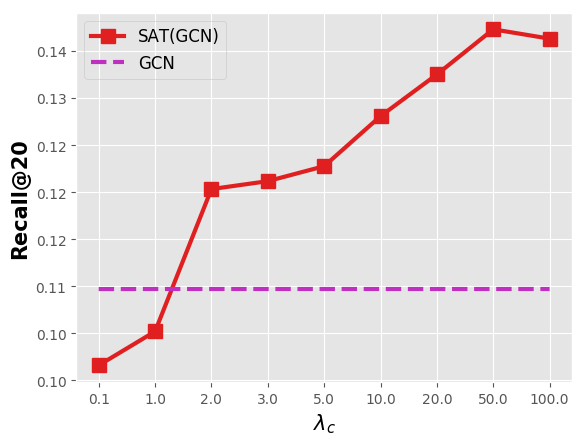}
\caption*{(e) Recall - Citeseer}
\end{minipage}
\begin{minipage}[t]{0.3\textwidth}
\centering
\includegraphics[width=\textwidth]{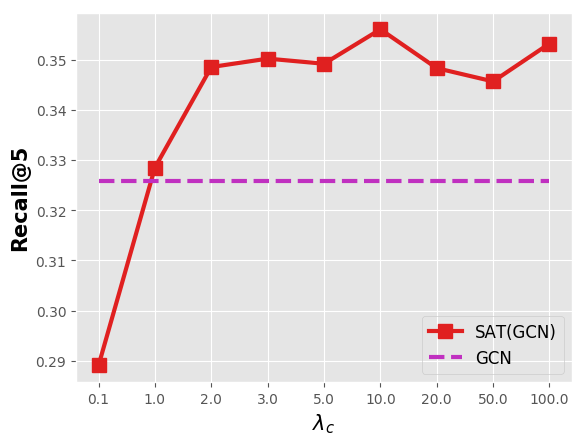}
\caption*{(f) Recall - Steam}
\end{minipage}\\
\begin{minipage}[t]{0.3\textwidth}
\centering
\includegraphics[width=\textwidth]{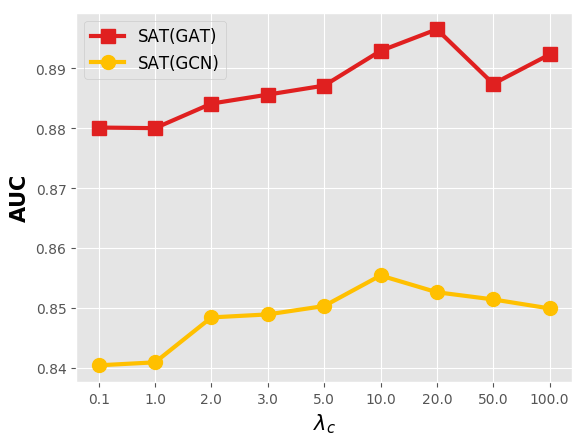}
\caption*{(g) AUC - Cora}
\end{minipage}
\begin{minipage}[t]{0.3\textwidth}
\centering
\includegraphics[width=\textwidth]{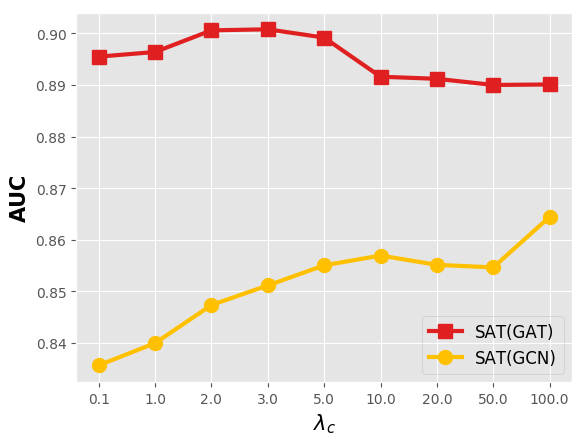}
\caption*{(h) AUC - Citeseer}
\end{minipage}
\begin{minipage}[t]{0.3\textwidth}
\centering
\includegraphics[width=\textwidth]{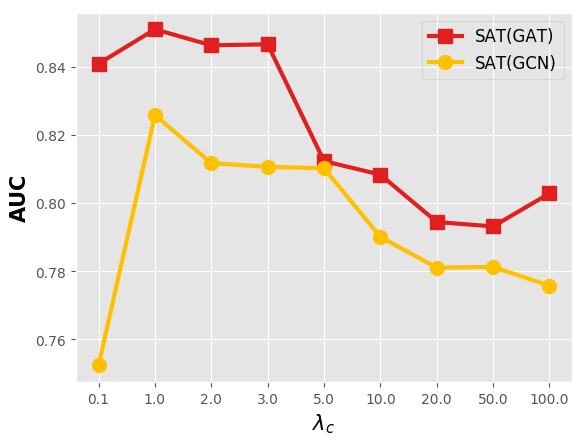}
\caption*{(i) AUC - Pubmed}
\end{minipage}
\caption{SAT(GCN) performance with different $\lambda_\mathrm{c}$ on both the node classification with ``A+X'' setting and profiling task. (a-c) means the result for node classification with ``A+X'' setting on Cora, Citeseer and Pubmed, respectively. The dotted line with "only A" represents that only the structural information is used, in which GCN as the classifier. (d-f) indicates the result for profiling on Cora, Citeseer and Steam, respectively. The dotted line with "GCN" means we use the GCN as the attribute completion method. (g-i) shows the link prediction performance of SAT(GCN) and SAT(GAT) with different $\lambda_{c}$.}
\label{figure:lambda_c}
\end{figure*}
\subsection{Learning Process Visualization}
In order to better understand the learning process of SAT, we plot learning curves including the train reconstruction loss, train GAN loss, validation metric, and MMD distance along the training epochs. The results are shown in Figure~\ref{figure:Train_process}.

From Figure~\ref{figure:Train_process}, we can summarize that that both the train reconstruction loss in (a)(b) and train GAN loss in (c) converge in the training process. And the validation Recall@10 in (d) increases step by step and finally converges at around $800th$ epoch. The MMD distance between learned aggregated distribution $q(z)$ and Gaussian prior $p(z)$ in (e) decreases step by step, which shows $q(z)$ matches the whole distribution of $p(z)$ successfully. 

Furthermore, since SAT involves distribution matching between latent codes $z_{x}$ encoded from attributes and $z_{a}$ encoded from structures, thus it is necessary to see whether our method matches them as what we expected. We show the train and validation MMD distance between $z_{x}$ and $z_{a}$ in Figure~\ref{figure:Train_process} (f). In this figure, the MMD distance between $z_{x}$ and $z_{a}$ is minimized gradually. This also keeps consistency with our \emph{shared-latent space} assumption and further benefits the joint distribution modeling on attribute-missing graphs.

\subsection{Hyper-parameter $\lambda_\mathrm{c}$}
In SAT, we introduce $\lambda_\mathrm{c}$ to weight the cross-reconstruction stream in the objective function. It is desirable to see how our method responds to this hyper-parameter. Intuitively, we conduct an experiment about the \emph{node attribute completion} and link prediction performance with different $\lambda_\mathrm{c}=[0.1,1.0,2.0,3.0,5.0,10.0,20.0,50.0,100.0]$. The results are shown in Figure~\ref{figure:lambda_c}. 

Figure~\ref{figure:lambda_c} (a-c) shows the results of node classification with ``A+X'' setting on Cora, Citeseer and Pubmed. Figure~\ref{figure:lambda_c} (d-f) indicates the results of profiling on Cora, Citeseer and Steam. From Figure~\ref{figure:lambda_c} (a-f), we can see that we need a large $\lambda_\mathrm{c}$ to restore high-quality node attributes. $\lambda_\mathrm{c}$ is vital for \emph{node attribute completion} since we rely on the cross-reconstruction stream to restore node attributes.

For link prediction in Figure~\ref{figure:lambda_c} (g-i), it seems that link prediction performance is more robust with $\lambda_{c}$ compared to the \emph{node attribute completion} task. For example, as shown in Figure~\ref{figure:lambda_c} (g) and (a), the relative change of AUC is 0.90$\sim$0.84 while that of accuracy is 0.84$\sim$0.72. This is mainly because \emph{node attribute completion} is a more difficult task since it generates high-dimensional data, which requires more fine-grained restoration and less disturbance. While for link prediction, it is conducted by the inner product between two low-dimensional embeddings of nodes, which could stand more disturbance from $\lambda_{c}$.

\begin{figure*}[h!]
\centering
\begin{minipage}[t]{0.32\textwidth}
\centering
\includegraphics[width=\textwidth]{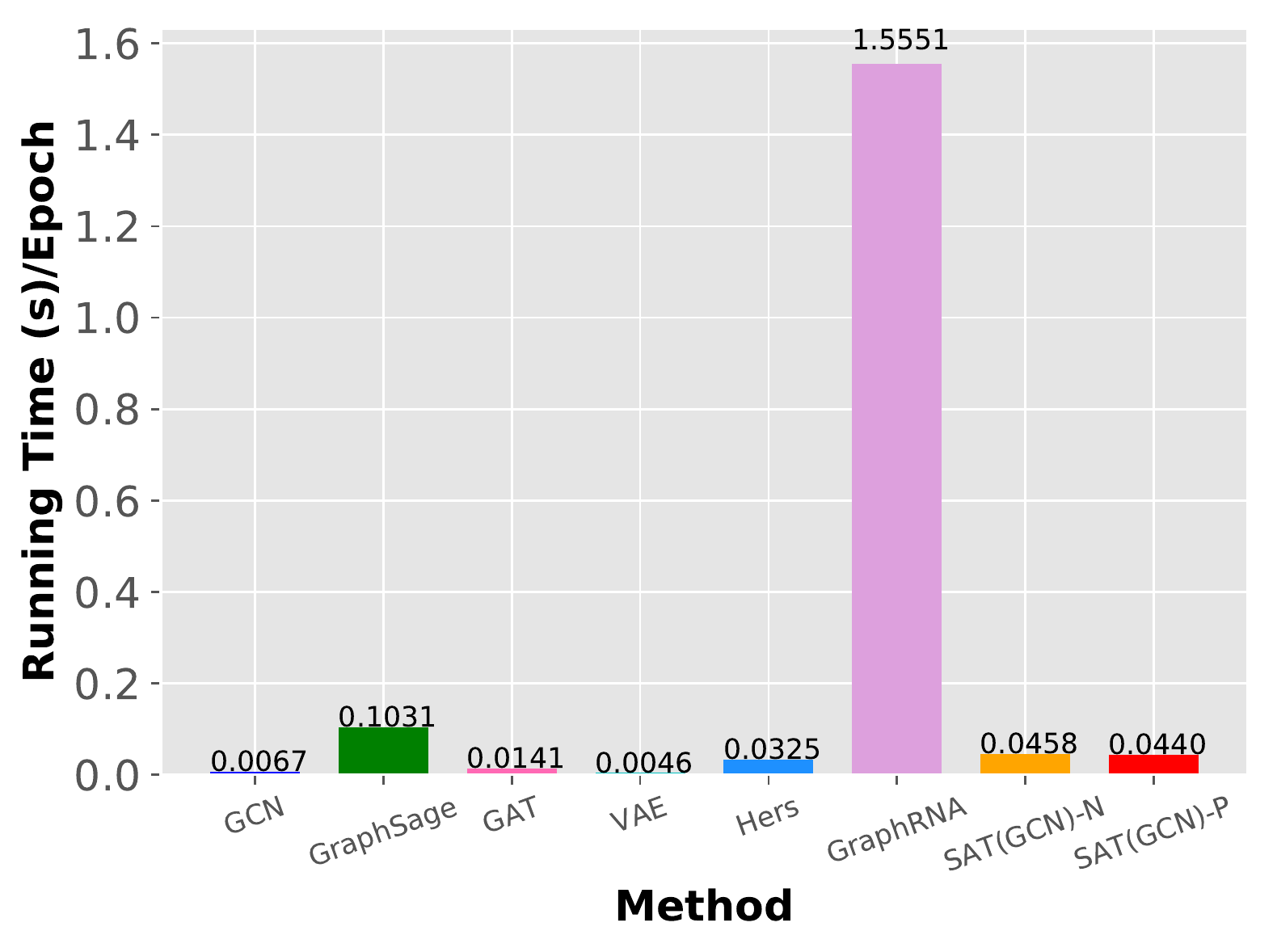}
\vspace{-17pt}
\caption*{(a) Cora}
\end{minipage}
\begin{minipage}[t]{0.32\textwidth}
\centering
\includegraphics[width=\textwidth]{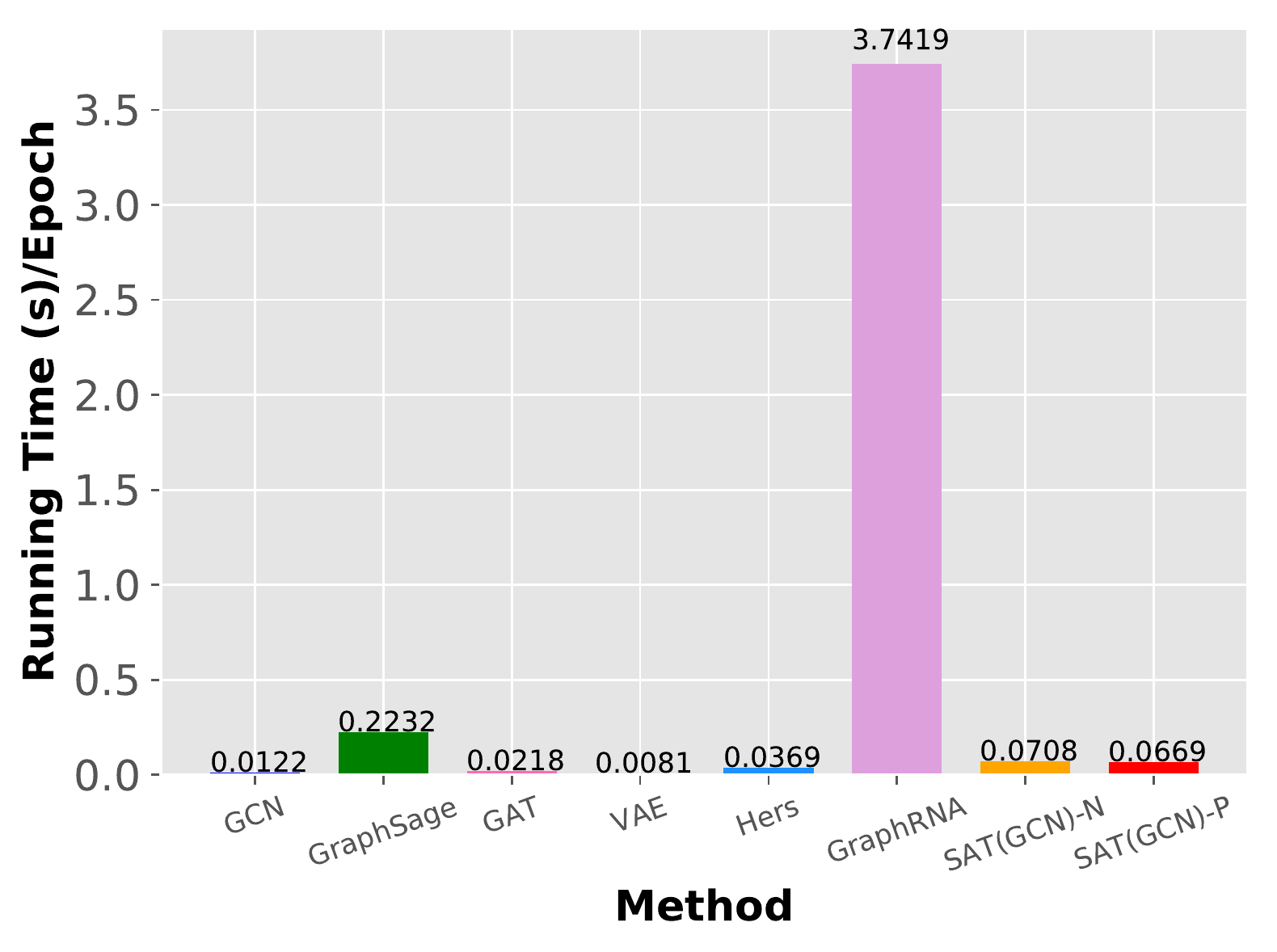}
\vspace{-17pt}
\caption*{(b) Citeseer}
\end{minipage}
\begin{minipage}[t]{0.32\textwidth}
\centering
\includegraphics[width=\textwidth]{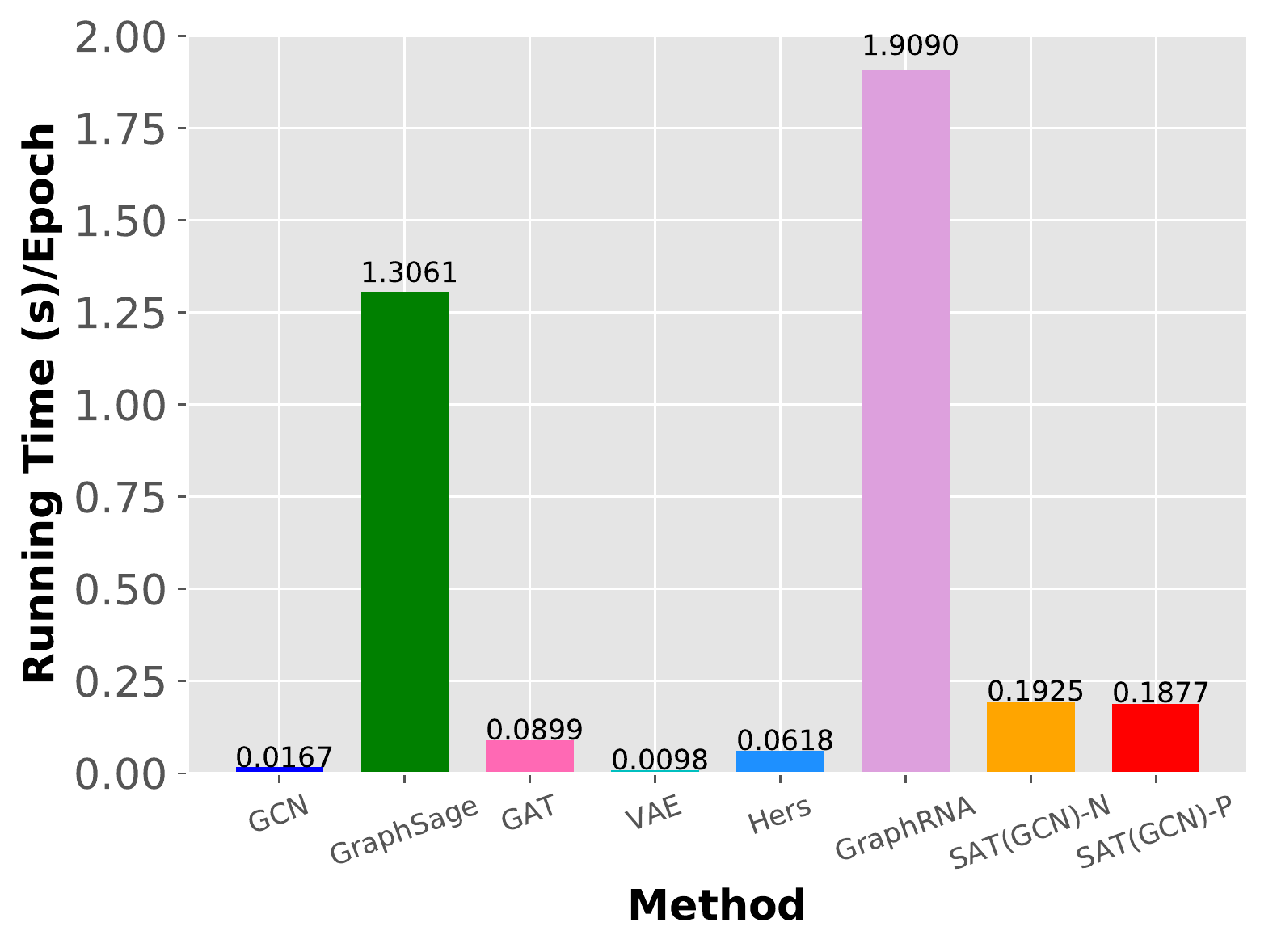}
\vspace{-17pt}
\caption*{(c) Steam}
\end{minipage} 
\caption{The empirical running time in each epoch of different methods. In this figure, SAT(GCN)-N indicates the non-parallel SAT(GCN), and SAT(GCN)-P means the parallel one.}
\label{figure:running_time}
\end{figure*}
\subsection{Empirical Running Time Analysis}
To investigate the time complexity, we conduct an experiment to compare the empirical running time of each epoch for different methods. Since SAT has parallel versions, we thus denote SAT-N as the non-parallel one, SAT-P as the parallel one. We conduct the experiments 10 times on the same machine with one Nvidia-TitanX GPU. The mean value of running time per epoch is reported in Figure~\ref{figure:running_time}. 

From Figure~\ref{figure:running_time}, we can see that: (1) GraphRNA costs the most time because it involves the LSTM for feature encoding. Although SAT(GCN) involves a two-level distribution matching, it takes less time than GraphRNA. Taking the previous experimental results into consideration, we see that the proposed SAT generally achieves better performance with less time complexity compared to GraphRNA. (2) SAT(GCN)-N and SAT(GCN)-P have similar time consumption. This indicates that the time cost of MLP encoding is trivial compared to the GNN encoding. In practice, we can use the non-parallel SAT instead of the parallel one.

\section{Conclusion and Future Work}
In this paper, we explore the learning problems on attribute-missing graphs and make a \emph{shared-latent space} assumption on graphs. Based on the assumption, we develop a novel distribution matching based GNN called structure-attribute transformer (SAT). SAT can not only handle the link prediction task but also the introduced \emph{node attribute completion} task on graphs. Furthermore, for the \emph{node attribute completion} task, we introduce practical measures including both node classification in the \emph{node level} and profiling in the \emph{attribute level} to evaluate the quality of restored node attributes. Empirical results validate the superiority of our method on both \emph{node attribute completion} and link prediction task. 

Learning on attribute-missing graphs is still an open problem and many topics could be studied from both the methodology and application aspect. For example, under our SAT framework, more efficient distribution matching methods for this problem could be investigated. There are also various real-world applications could be explored such as fraud detection in social networks, author description generation in co-author networks and image caption generation. Furthermore, more complex graph data such as heterogeneous attribute-missing graphs could also be an interesting problem. These will be studied in future works.

\bibliographystyle{abbrv}
\bibliography{cite}
\ifCLASSOPTIONcaptionsoff
\newpage
\fi

\vspace{0cm}\begin{IEEEbiography}[{\includegraphics[width=1in,height=1.25in,clip,keepaspectratio]{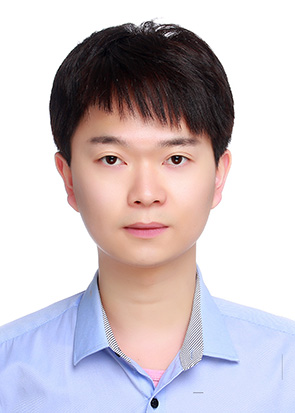}}]{Xu Chen} received the B.S. degree in electronics engineering from Xi Dian University in 2016. He is working toward the Ph.D. degree at Cooperative Meidianet Innovation Center in Shanghai Jiao Tong University since 2016. He is now a dual Ph.D. student of Shanghai Jiao Tong University and University of Technology, Sydney. His research interests include machine learning, graph representation learning, recommendation systems, and computer vision. \end{IEEEbiography}

\vspace{0cm}\begin{IEEEbiography}[{\includegraphics[width=1in,height=1.25in,clip,keepaspectratio]{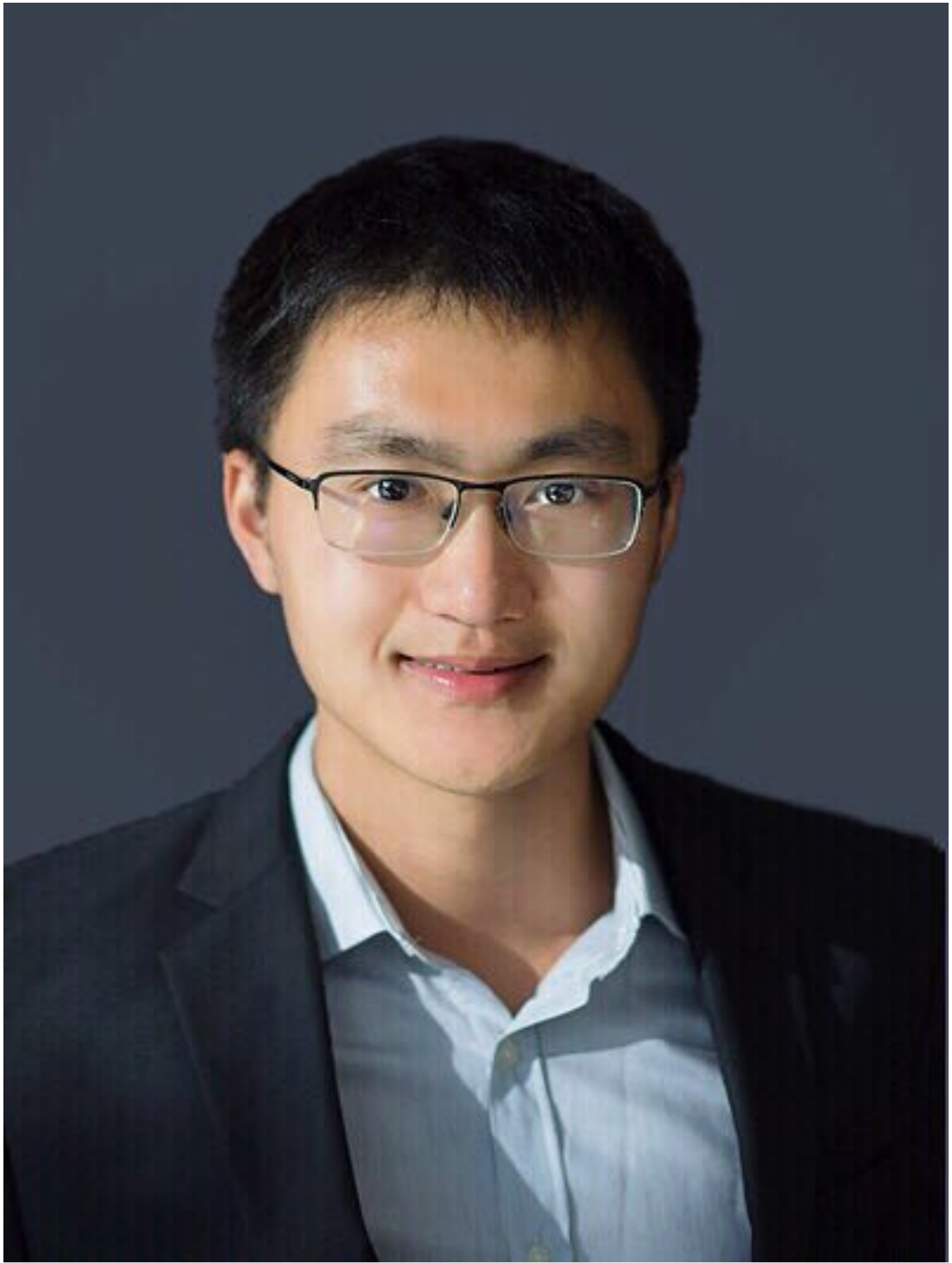}}]{Siheng Chen} is a research scientist at Mitsubishi Electric Research Laboratories (MERL). Before that, he was an autonomy engineer at Uber Advanced Technologies Group, working on the perception and prediction systems of self-driving cars. Before joining Uber, he was a postdoctoral research associate at Carnegie Mellon University. Chen received the doctorate in Electrical and Computer Engineering from Carnegie Mellon University in 2016, where he also received two masters degrees in Electrical and Computer Engineering and Machine Learning, respectively. He received his bachelor's degree in Electronics Engineering in 2011 from Beijing Institute of Technology, China. Chen was the recipient of the 2018 IEEE Signal Processing Society Young Author Best Paper Award. His coauthored paper received the Best Student Paper Award at IEEE GlobalSIP 2018. His research interests include graph signal processing, graph neural networks and 3D computer vision. \end{IEEEbiography}

\vspace{0cm}\begin{IEEEbiography}[{\includegraphics[width=1in,height=1.25in,clip,keepaspectratio]{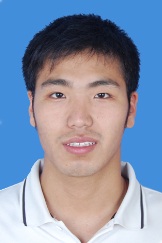}}]{Jiangchao Yao} received the B.S. degree in information engineering from the South China University
of Technology, Guangzhou, China, in 2013. He
is currently pursuing the dual Ph.D. degree under
the supervision of Y. Zhang with Shanghai Jiao
Tong University and under the supervision of Ivor W
Tsang with the University of Technology Sydney.
His research interests include representation learning
and multimedia analysis. \end{IEEEbiography}

\vspace{0cm}\begin{IEEEbiography}[{\includegraphics[width=1in,height=1.25in,clip,keepaspectratio]{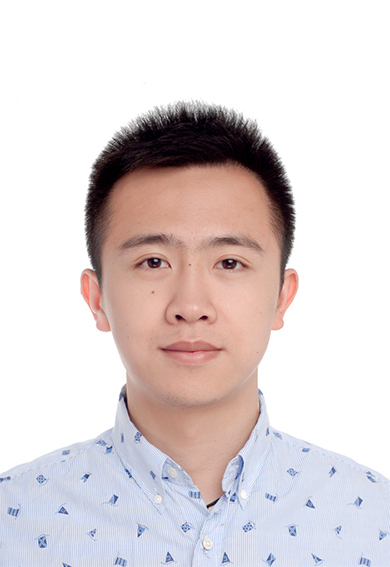}}]{Huangjie Zheng} received the BS and MS degrees from Shanghai Jiao Tong University in 2019 and is currently working toward the PhD degree at the University of Texas at Austin. His research interests include machine learning and Bayesian deep learning, with specific interests in probabilistic modeling and deep generative models with application to computer vision. \end{IEEEbiography}

\vspace{0cm}\begin{IEEEbiography}[{\includegraphics[width=1in,height=1.25in,clip,keepaspectratio]{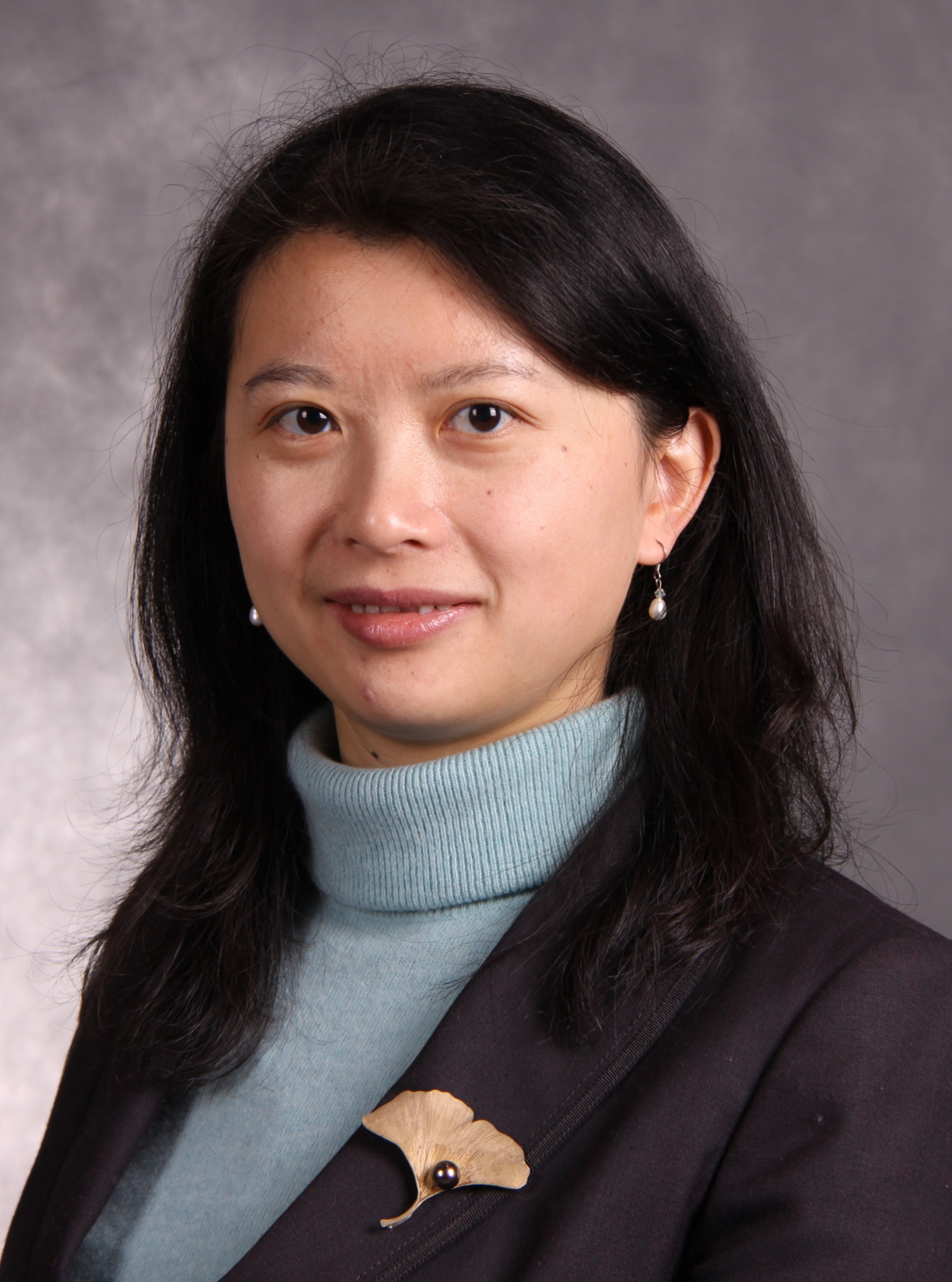}}]{Ya Zhang} received the B.S. degree from Tsinghua University and the Ph.D. degree in information sciences and technology from the Pennsylvania State University. Since March 2010, she has been a professor with Cooperative Medianet Innovation Center, Shanghai Jiao Tong University. Prior to that, she worked with Lawrence Berkeley National Laboratory, University of Kansas, and Yahoo! Labs. Her research interest is mainly on data mining and machine learning, with applications to information retrieval, web mining, and multimedia analysis. She is a member of the IEEE. \end{IEEEbiography}

\vspace{0cm}\begin{IEEEbiography}[{\includegraphics[width=1in,height=1.25in,clip,keepaspectratio]{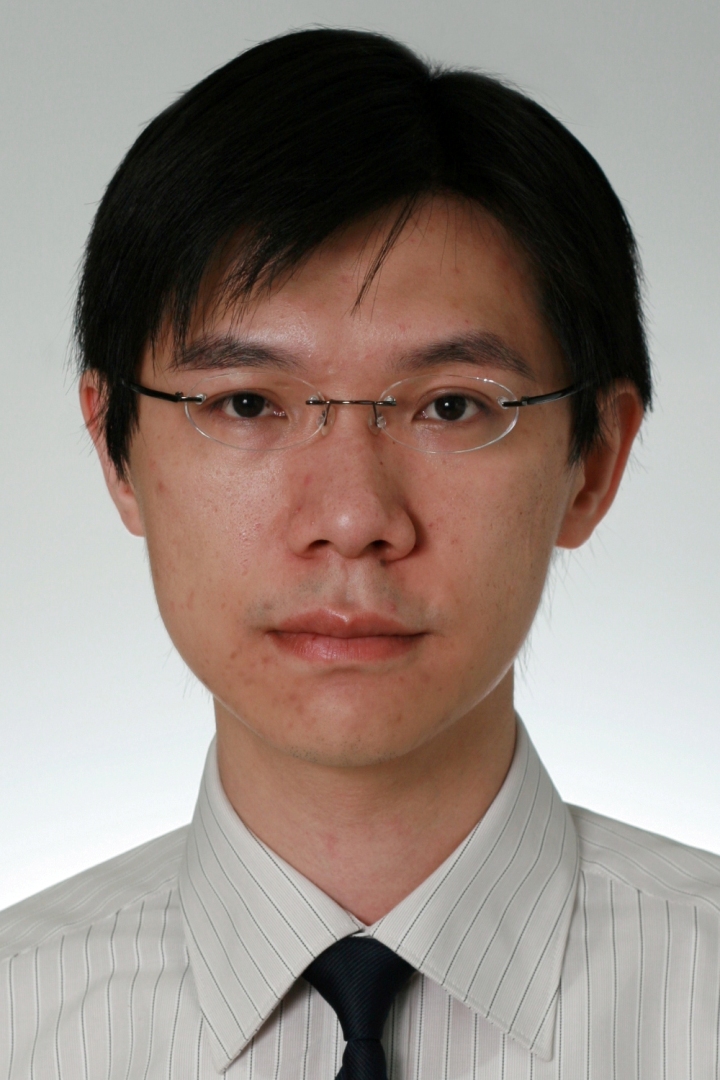}}]{Ivor W.Tsang} is Professor of Artificial Intelligence with the University of Technology Sydney. He is also the Research Director of the Australian Artificial Intelligence Institute. His research interests include transfer learning, deep generative models, and big data analytics.  In 2013, Prof Tsang received his prestigious ARC Future Fellowship for his research regarding Machine Learning on Big Data. In 2019, his JMLR paper titled "Towards ultrahigh dimensional feature selection for big data" received the International Consortium of Chinese Mathematicians Best Paper Award. In 2020, Prof Tsang was recognized as the AI 2000 AAAI/IJCAI Most Influential Scholar in Australia for his outstanding contributions to the field of AAAI/IJCAI between 2009 and 2019. His research on transfer learning granted him the Best Student Paper Award at CVPR 2010 and the 2014 IEEE TMM Prize Paper Award. In addition, he received the IEEE TNN Outstanding 2004 Paper Award in 2007. He serves as a Senior Area Chair/Area Chair for NeurIPS, ICML, AISTATS, AAAI and IJCAI,  and the Editorial Board for JMLR, MLJ, and IEEE TPAMI. \end{IEEEbiography}

\end{document}